\documentclass{article}

\usepackage[nonatbib,preprint]{neurips_2020}

\usepackage[numbers]{natbib}
\usepackage{color}
\usepackage{algorithm}
\usepackage[noend]{algpseudocode}
\usepackage[utf8]{inputenc} 
\usepackage[T1]{fontenc}    
\usepackage{hyperref}       
\usepackage{url}            
\usepackage{booktabs}       
\usepackage{amsfonts}       
\usepackage{nicefrac}       
\usepackage{microtype}      
\usepackage[linewidth=1pt]{mdframed}
\usepackage{tikz-dependency}
\usepackage{amsmath,amsfonts,bm}
\usepackage{graphicx}
\usepackage{algorithmicx}
\usepackage[noend]{algpseudocode}
\usepackage{booktabs}
\usepackage[percent]{overpic}
\usepackage{subcaption}
\usepackage{siunitx}
\usepackage{wrapfig}
\usepackage{listings}
\usepackage{multirow}
\usepackage[page]{appendix}

\usepackage{hyperref}
\usepackage{url}
\NewEnviron{elaboration}{
\par
\begin{tikzpicture}
\node[rectangle,minimum width=\textwidth] (m) {\begin{minipage}{.98\textwidth}\BODY\end{minipage}};
\draw[dashed] (m.south west) rectangle (m.north east);
\end{tikzpicture}
}
\usepackage{enumitem}

\usepackage[export]{adjustbox}

\newcommand\qbert{Q*BERT}

\title{How to Avoid Being Eaten by a Grue: \\ Structured Exploration Strategies for Textual Worlds}

\author{
Prithviraj Ammanabrolu $^{\dagger}$ \:\:\:\: Ethan Tien $^{\dagger}$ \:\:\:\: Matthew Hausknecht$^{\ddagger}$ \:\:\:\: Mark O. Riedl $^{\dagger}$ \\
$^{\dagger}$Georgia Institute of Technology \:\:\:\:  $^{\ddagger}$Microsoft Research\\
\texttt{raj.ammanabrolu@gatech.edu}
}

\begin{document}

\maketitle

\begin{abstract}
Text-based games are long puzzles or quests, characterized by a sequence of sparse and potentially deceptive rewards.
They provide an ideal platform to develop agents that {\em perceive} and {\em act upon} the world using a combinatorially sized natural language state-action space.
Standard Reinforcement Learning agents are poorly equipped to effectively explore such spaces and often struggle to overcome bottlenecks---states that agents are unable to pass through simply because they do not see the right action sequence enough times to be sufficiently reinforced.
We introduce \qbert{}~\footnote{Code can be found here \url{https://github.com/rajammanabrolu/Q-BERT}}, an agent that learns to build a knowledge graph of the world by answering questions, which leads to greater sample efficiency.
To overcome bottlenecks, we further introduce MC!\qbert{} an agent that uses an knowledge-graph-based intrinsic motivation to detect bottlenecks and a novel exploration strategy to efficiently learn a chain of policy modules to overcome them.
We present an ablation study and results demonstrating how our method outperforms the current state-of-the-art on nine text games, including the popular game, {\em Zork}, where, for the first time, a learning agent gets past the bottleneck where the player is eaten by a Grue.
\end{abstract}

\section{Introduction}

\begin{wrapfigure}[24]{r}{.5\textwidth}
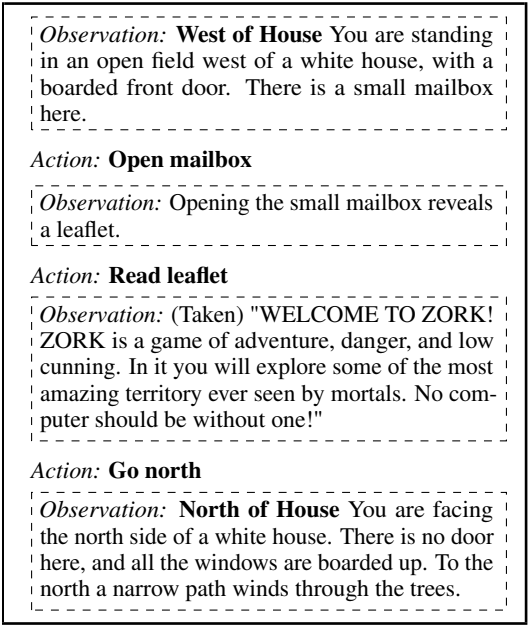

\footnotesize
\begin{mdframed}
\begin{elaboration}
  \parbox{.99\textwidth}{
\emph{Observation:} \textbf{West of House} You are standing in an open field west of a white house, with a boarded front door. There is a small mailbox here.
}
\end{elaboration}
\begin{flushleft}
\emph{Action:} \textbf{Open mailbox}
\end{flushleft}
\begin{elaboration}
  \noindent\parbox{.99\textwidth}{
\emph{Observation:} Opening the small mailbox reveals a leaflet.
}
\end{elaboration}
\begin{flushleft}
\emph{Action:} \textbf{Read leaflet}
\end{flushleft}
\begin{elaboration}
  \noindent\parbox{.99\textwidth}{
\emph{Observation:} (Taken) "WELCOME TO ZORK!
ZORK is a game of adventure, danger, and low cunning. In it you will explore some of the most amazing territory ever seen by mortals. No computer should be without one!"
}
\end{elaboration}
\begin{flushleft}
\emph{Action:} \textbf{Go north}
\end{flushleft}
\begin{elaboration}
  \parbox{.99\textwidth}{
\emph{Observation:} \textbf{North of House}
You are facing the north side of a white house. There is no door here, and all the windows are boarded up. To the north a narrow path winds through the trees.
}
\end{elaboration}
\end{mdframed}
\caption{Excerpt from {\em Zork1}.}
\label{fig:zorkexcerpt}
\end{wrapfigure}

Text-adventure---or interaction fiction---games are simulations featuring language-based state and action spaces.
An example of a one turn agent interaction in the popular text-game {\em Zork1}~\citep{zork} can be seen in Fig.~\ref{fig:zorkexcerpt}.
Prior works have focused on a few challenges that are inherent to this medium:
(1)~{\em Partial observability} the agent must reason about the world solely through incomplete textual descriptions~\citep{narasimhan15,cote18,ammanabrolu}.
(2)~{\em Commonsense reasoning} to enable the agent to more intelligently interact with objects in its surroundings~\citep{fulda17,jonmay,adolphs2019ledeepchef,ammanabrolutransfer}.
(3)~{\em A combinatorial state-action space} wherein 
most games have action spaces exceeding a billion possible actions per step; for example the game {\em Zork1} has $\num{1.64e14}$ possible actions at every step~\citep{jericho,Ammanabrolu2020Graph}.
Despite these challenges, modern text-adventure agents such as KG-A2C~\cite{Ammanabrolu2020Graph}, TDQN~\cite{jericho}, and DRRN~\cite{he16} have relied on surprisingly simple exploration strategies such as $\epsilon$-greedy or sampling from the distribution of possible actions.


In this paper, we focus on a particular type of exploration problem: that of detecting and overcoming bottleneck states.
Most text-adventure games have relatively linear plots in which players must solve a sequence of puzzles to advance the story and gain score.
To solve these puzzles, players have freedom to a explore previously unlocked areas of the game, collect clues, and acquire tools needed to solve the next puzzle and unlock the next portion of the game.
From a Reinforcement Learning perspective, 
These puzzles can be viewed as bottlenecks that act as partitions between different regions of the state space.
We contend that existing Reinforcement Learning agents are unaware of such latent structure and are thus poorly equipped for solving these types of problems.
We present techniques for automatically detecting bottlenecks and efficiently learning policies that take advantage of the natural partitions in the state space.

Overcoming bottlenecks is not as simple as selecting the correct action from the bottleneck state. 
Most bottlenecks have 
long-range
dependencies that must first be satisfied:
{\em Zork1} for instance features a bottleneck in which the agent must pass through the unlit {\em Cellar} where a monster known as a Grue lurks, ready to eat unsuspecting players who enter without a light source.
To pass this bottleneck the player must have previously acquired and lit the latern.
Other bottlenecks don't rely on inventory items and instead require the player to have satisfied an external condition such as visiting the reservoir control to drain water from a submerged room before being able to visit it.
In both cases, the actions that fulfill dependencies of the bottleneck, e.g. acquiring the lantern or draining the room, are not rewarded by the game. 
Thus agents must correctly satisfy all \emph{latent} dependencies, most of which are unrewarded, then take the right action from the correct location to overcome such bottlenecks.
Consequently,
most existing agents---regardless of whether they use a reduced action space~\citep{zahavy18,jonmay} or the full space~\citep{jericho,Ammanabrolu2020Graph}---have failed to consistently clear these bottlenecks.

\begin{figure}[t]
    \centering
    \includegraphics[width=\linewidth]{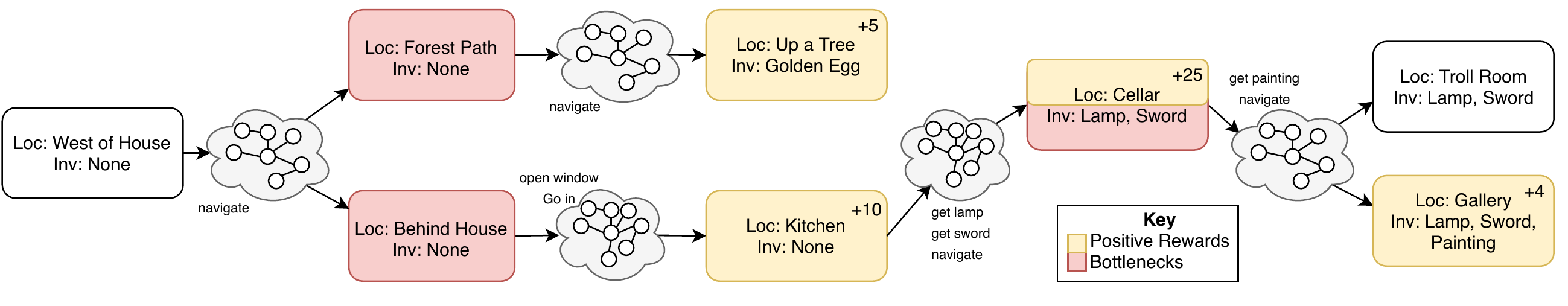}
    \caption{Portion of the {\em Zork1} quest structure visualized as a directed acyclic graph. 
    Each node represents a state; 
    clouds represent areas of high branching factor
    with labels indicating some of the actions that must be performed to progress
    }
\label{fig:dag}
\end{figure}

To better understand how to design algorithms that pass these bottlenecks, we first need to gain a sense for what they are.
We observe
that quests in text games---and any such sequential decision making problem requiring long term dependencies---can be modeled in the form of a dependency graph.
These dependency graphs are directed acyclic graphs (DAGs) where the vertices indicate either rewards that can be collected or dependencies that must be met to progress.
In text-adventure games the dependencies are of two types:
items that must be collected for future use, and locations that must be visited.
An example of such a graph for the game of {\em Zork1} can found in Fig.~\ref{fig:dag}.
More formally, bottleneck states are vertices in the dependency graph that, when the graph is laid out topographically, are (a)~the only state on a level, and (b)~there is another state at a higher level with non-zero reward.
Bottlenecks can be mathematically expressed as follows: let $\mathcal{D}=\langle V, E\rangle$ be the directed acyclic dependency graph for a particular game where each vertex is tuple $v=\langle s_l, s_i, r(s)\rangle$ containing information on some state $s$ such that $s_l$ are location dependencies, $s_i$ are inventory dependencies, and $r(s)$ is the reward associated with the state.
There is a directed edge $e\in E$ between any two vertices such that the originating state meets the requirements $s_l$ and $s_i$ of the terminating vertex.
$\mathcal{D}$ can be topologically sorted into levels $L=\{l_1,...,l_n\}$ where each level represents a set of game states that are not dependant on each other.
We formulate the set of all bottleneck states in the game: 
\begin{equation}
\label{eq:bottleneck}
    \mathcal{B}=\{b: (|l_i|=1, b\in l_i, V) \land (\exists s \in l_j \text{ s.t. } (j > i \land r(s) \neq 0))\}
\end{equation}

This reads as the set of all states that that belong to a level with only one vertex and that there exists some state with a non-zero reward that depends on it.
Intuitively, regardless of the path taken to get to a bottleneck state, any agent must pass it in order to continue collecting future rewards.
{\em Behind House} is an example of a bottleneck state as seen in Fig.~\ref{fig:dag}.
The branching factor before and after this state is high but it is the only state through which one can enter the {\em Kitchen} through the window.


In this paper, 
we introduce \qbert{}, a deep reinforcement learning agent that plays text games by building a knowledge graph of the world and answering questions about it. 
Knowledge graph state representations have been shown to alleviate other challenges associated with text games such as partial-observability~\citep{ammanabrolu,yin2019learn,Ammanabrolu2020Graph,ammanabrolu2020bringing,adhikari2020learning,murugesan2020enhancing}.
We introduce the {\em Jericho-QA} dataset, for question-answering in text-game-like environments, and show that our novel question-answering-based graph construction procedure improves sample efficiency but not asymptotic performance.
In order to improve performance and pass through bottlenecks, we extend \qbert{} with a novel exploration strategy that uses intrinsic motivation based on the knowledge graph to alleviate the sparse, deceptive reward problem.
Our exploration strategy first detects bottlenecks and then modularly chains policies that go from one bottleneck to another.
We call this combined system
MC!\qbert{}.
These two enhancements form the two core contributions of this paper.
We evaluate \qbert{}, MC!\qbert{}, and ablations of both on a set of nine text games.
We further compare our technique to alternative exploration methods such as Go Explore~\citep{ecoffet19}; our full technique achieves state-of-the-art performance on eight out of nine games.

\section{Related Work and Background}

We use the definition of text-adventure games as seen in~\citet{cote18} and~\citet{jericho}.
These games are partially observable Markov decision processes (POMDPs), represented as a 7-tuple of $\langle S,T,A,\Omega , O,R, \gamma\rangle$ representing the set of environment states, {\em mostly deterministic conditional transition probabilities between states}, the vocabulary or words used to compose text commands, observations returned by the game, observation conditional probabilities, reward function, and the discount factor respectively.
%
%
LSTM-DQN~\citep{narasimhan15} and Action Elimination DQN~\citep{zahavy18} operate on a reduced action space of the order of $10^2$ actions per step by considering either verb-noun pairs or by using a walkthrough of the game respectively.
The agents learn how to produce Q-value estimates that maximize long term expected reward.
The DRRN algorithm for choice-based games~\citep{he16,zelinka18} estimates Q-values for a particular action from a particular state. 
\citet{fulda17} try to use word embeddings specifically in an attempt to model affordances for items in these games, learning how to interact with them.

There have been a couple of works detailing potential methods of exploration in this domain.
\citet{jain2019algorithmic} extend consistent Q-learning~\citep{bellemare2016increasing} to text-games, focusing on taking into account historical context.
In terms of exploration strategies, \citet{yuan18} detail how counting the number of unique states visited improves generalization in unseen games.
\citet{cote18} introduce TextWorld, a framework for procedurally generating parser-based games via a grammar, allowing a user to control the difficulty of a generated game.
\citet{urbanek2019light} introduce LIGHT, a dataset of crowdsourced text-adventure game dialogs focusing on giving collaborative agents the ability to generate contextually relevant dialog and emotes.
\citet{jericho} introduce Jericho, a framework for interacting with text-games, in addition to a series of baseline reinforcement learning agents.
\citet{yuan2019qait} introduce the concept of interactive question-answering in the form of QAit---modeling QA tasks in TextWorld.

\citet{ammanabrolu} introduce the KG-DQN, using knowledge graphs as states spaces for text-game agents and \citet{ammanabrolutransfer} extend it to enable transfer of knowledge between games.
\citet{Ammanabrolu2020Graph} showcase the KG-A2C, for the first time tackling the fully combinatorial action space and presenting state-of-the-art results on many man-made text games.
In a similar vein, \citet{adhikari2020learning} present the Graph-Aided Transformer Agent (GATA) which learns to construct a knowledge graph during game play and improves zero-shot generalization on procedurally generated TextWorld games.


\section{\qbert}
\label{sec:qbertall}

\begin{figure}
    \centering
    \includegraphics[width=.9\linewidth]{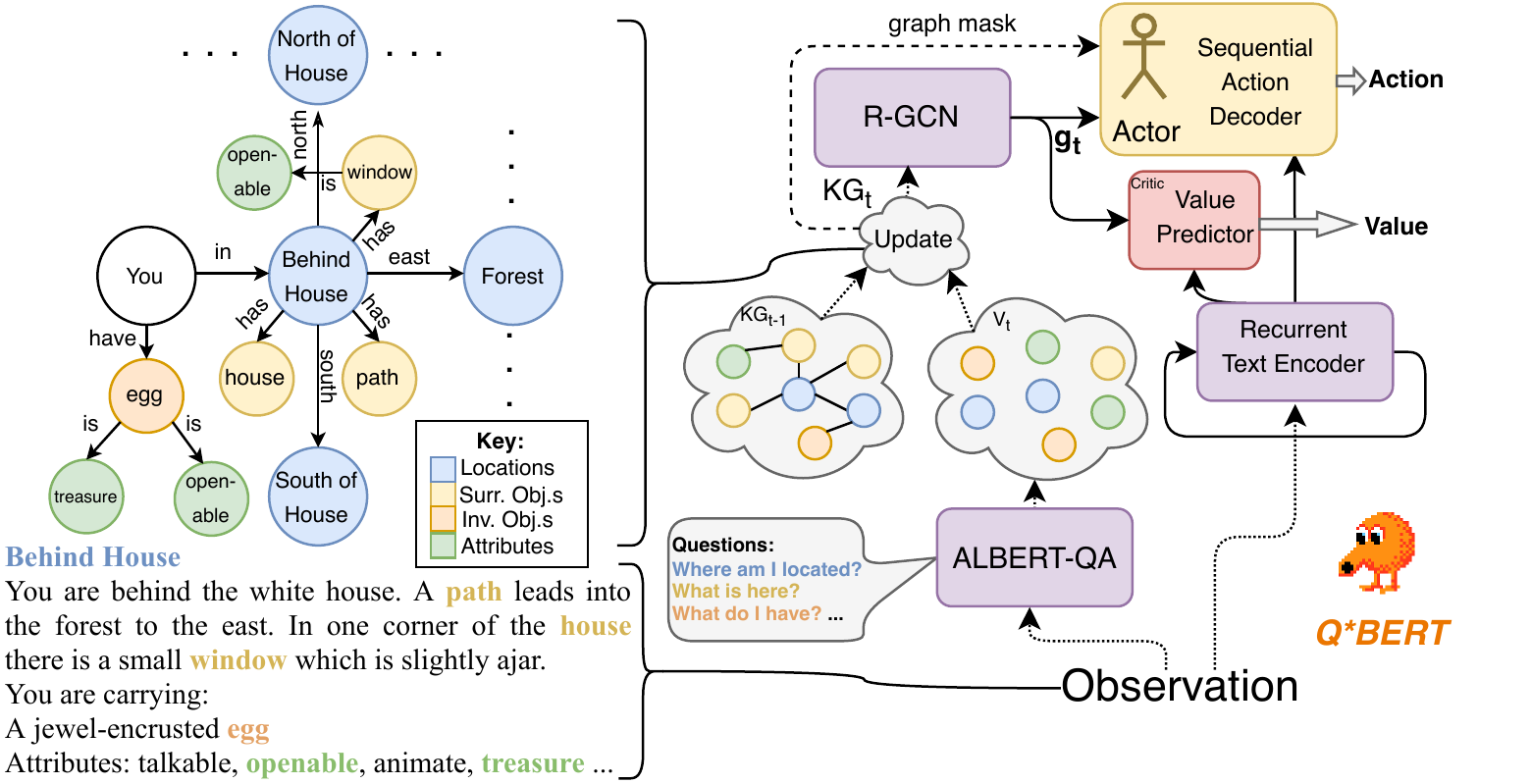}
    \caption{One-step knowledge graph extraction in the Jericho-QA format, and overall \qbert{} architecture at time step $t$. At each step the ALBERT-QA model extracts a relevant highlighted entity set $V_t$ by answering questions based on the observation, which is used to update the knowledge graph.}
    \label{fig:architecture}
\end{figure}

This section presents the base reinforcement learning algorithm we introduce, which we call \qbert{}.
\qbert{} is based on KG-A2C~\citep{Ammanabrolu2020Graph}; it
uses a knowledge-graph to represent it's understanding of the world state.
A knowledge graph is a set of relations $\langle s, r, o\rangle$ such that $s$ is a subject, $r$ is a relation, and $o$ is an object.
See Figure~\ref{fig:architecture} (left) for an example fragment of a knowledge graph for a text-adventure game.
Instead of using relation extraction rules, \qbert{} uses a variant of the BERT~\citep{devlin18} natural language transformer to answer questions about the current state text description and populate the knowledge graph from the answers.


\paragraph{Knowledge Graph State Representation}

\citet{ammanabrolu} are the first to use question answering (QA) in text-game playing to pre-train a network to answer the question of ``What action best next to take?'' using game traces from an oracle agent capable of playing a game perfectly.
They pre-train an LSTM to predict the action based on a environment text description.
We build on this idea but instead treat the problem of constructing the knowledge graph as a question-answering task. 
The method first extracts a set of graph vertices $\mathcal{V}$ by asking a question-answering system relevant questions and then linking them together using a set of relations $\mathcal{R}$ to form a knowledge graph representing information the agent has learned about the world.
Examples of questions include: 
``What is my current location?'', 
``What objects are around me?'', and
''What am I carrying?'' to respectively extract information regarding the agent's current location, surrounding objects, inventory objects.
Further, we predict attributes for each object by asking the question ``What attributes does $x$ object have?''.
An example of the knowledge graph that can be extracted from description text and the overall architecture are shown in Fig.~\ref{fig:architecture}.


For question-answering, we use the pre-trained language model ALBERT~\citep{Lan2020ALBERT:},
a variant of BERT~\citep{devlin18} that is fine-tuned for question answering on the SQuAD~\citep{rajpurkar-etal-2016-squad} question-answering dataset.
We further fine-tune the ALBERT model on a dataset specific to the text-game domain.
This dataset, dubbed {\em Jericho-QA}, was created by making question answering pairs about text-games. 
{\em Jericho}~\citep{jericho}\footnote{\url{https://github.com/microsoft/jericho}} is a framework for reinforcement learning in text-games.
Using Jericho we construct a question-answering corpus for fine-tuning ALBERT as follows.
For each game in Jericho, we use an oracle---an agent capable of playing the game perfectly---and a random exploration agent to gather ground truth state information about locations, objects, and attributes.
These agents are designed to  extract this information directly from the game engine, which is otherwise off-limits when \qbert{} is trained.
From this ground truth, we construct pairs of questions in the form that Q*BERT will ask as it encounters environment description text, and the corresponding answers. 
These question-answer pairs are used to fine-tune the Q/A model and the ground truth data is discarded.
No data from games we test on is used during ALBERT fine-tuning.
Additional details regarding Jericho-QA, graph update rules, and \qbert{} can be found in Appendix~\ref{app:qbert}.

In a text-game the observation is a textual description of the environment. 
For every observation received, \qbert{} produces a fixed set of questions.
The questions and the observation text are sent to the question-answering system.
Predicted answers are converted into $\langle s, r, o\rangle$ triples and added to the knowledge graph. 
The complete knowledge graph is the input into \qbert's neural architecture (training described below), which makes a prediction of the next action to take.

\paragraph{Action Space} 

Solving {\em Zork1}, the cannonical text-adventure game, requires the generation of actions consisting of up to five-words from a relatively modest vocabulary of 697 words recognized by the game’s parser.
This results in $\mathcal{O}(697^5)=\num{1.64e14}$ possible actions at every step.
\citet{jericho} propose a template-based action space in which the agent first selects a template, consisting of an action verb and preposition, and then filling that in with relevant entities $($e.g. $[get]$ \underline{\hspace*{.4cm}}$ [from] $\underline{\hspace*{.4cm}}$)$.
{\em Zork1} has 237 templates, each with up to two blanks, yielding a template-action space of size $\mathcal{O}(237 \times 697^2)=\num{1.15e8}$.
This space is still far larger than most used by previous approaches applying reinforcement learning to text-based games.
We use this template action space for all 
games.

\paragraph{Training}

%
At every step an observation consisting of several components is received: $o_t=(o_{t_{desc}},o_{t_{game}},o_{t_{inv}},a_{t-1})$ corresponding to the room description, game feedback, inventory, and previous action, and total score $R_t$.
The room description $o_{t_{desc}}$ is a textual description of the agent's location, obtained by executing the command ``look''.
The game feedback $o_{t_{game}}$ is the simulators response to the agent's previous action and consists of narrative and flavor text.
The inventory $o_{t_{inv}}$ and previous action $a_{t-1}$ components inform the agent about the contents of its inventory and the last action taken respectively.

Each of these components is processed using a GRU based encoder utilizing the hidden state from the previous step and combined to have a single observation embedding $\textbf{o}_t$.
At each step, we update our knowledge graph $G_t$ using $o_t$ as described in earlier in Section~\ref{sec:qbertall} and it is then embedded into a single vector $\mathbf{g_t}$.
This encoding is based on the R-GCN and is calculated as:
\begin{equation}
    \mathbf{g_t} = f\left(\mathbf{W_g}\sigma \left(\sum_{r \in \mathcal{R}}\sum_{j \in {\mathcal{N}_i}^r} \frac{1}{c_{i,r}}\mathbf{W_r}^{(l)}{\mathbf{h}_j}^{(l)} + \mathbf{W_0}^{(l)}{\mathbf{h}_i}^{(l)}\right) + \mathbf{b_g}\right)
\end{equation}
Where $\mathcal{R}$ is the set of relations, ${\mathcal{N}_i}^r$ is the 1-step neighborhood of a vertex $i$ with respect to relation $r$, $\mathbf{W_r}^{(l)}$ and ${\mathbf{h}_j}^{(l)}$ are the learnable convolutional filter weights with respect to relation $r$ and hidden state of a vertex $j$ in the last layer $l$ of the R-GCN respectively, $c_{i,r}$ is a normalization constant, and $\mathbf{W_g}$ and $\mathbf{b_g}$ the weights and biases of the output linear layer.
The full architecture can be found in Fig.~\ref{fig:architecture}.
The state representation consists only of the textual observations and knowledge graph.
Another key use of the knowledge graph, introduced as part of KG-A2C, is the {\em graph mask}, which restricts the possible set of entities that can be predicted to fill into the action templates at every step to those found in the agent's knowledge graph.
The rest of the training methodology is unchanged from \citet{Ammanabrolu2020Graph}, more details can be found in Appendix~\ref{app:qbert}.

\section{Structured Exploration}
This section describes an exploration method built on top of Q*BERT that first detects bottlenecks and then searches for ways to pass them, learning policies that take it from bottleneck to bottleneck.
This method of chaining policies and backtracking can be thought of in terms of {\em options}~\citep{options,sutton1999between},
where the agent decomposes the task of solving the text game into the sub-tasks, each of which has it's own policy. 
In our case, each sub-task delivers the agent to a bottleneck state.

\subsection{Bottleneck Detection using Intrinsic Motivation} 
Examples of some bottlenecks can be seen in Figure~\ref{fig:dag} based on our definition of a bottleneck in Eq.~\ref{eq:bottleneck}.
Inspired by \citet{mcgovern2001automatic}, we present an intuitive way of detecting these bottleneck states---or sub-tasks---in terms of whether or not the agent's ability to collect reward stagnates.
If the agent does not collect a new reward for a number of environment interactions---defined in terms of a {\em patience} parameter---then it is 
possible
that it is stuck due to a bottleneck state.
An issue with this method, however, is that the placement of rewards does not always correspond to an agent being stuck.
Complicating matters, rewards are sparse and often delayed; the agent not collecting a reward for a while might simply indicate that further exploration is required instead of truly being stuck.

To alleviate these issues, we define an {\em intrinsic motivation} for the agent that leverages the knowledge graph being built during exploration.
The motivation is for the agent to learn more information regarding the world and expand the size of its knowledge graph.
This provides us with a better indication of whether an agent is stuck or not---a stuck agent does not visit any new states, learns no new information about the world, and therefore does not expand its knowledge graph---leading to more effective bottleneck detection overall.
To prevent the agent from discovering reward loops based on knowledge graph changes, we formally define this reward in terms of new information learned.
\begin{equation}
    r_{\text{IM}_t} = \Delta (\mathcal{KG}_{\text{global}} - \mathcal{KG}_{t}) \;\; \text{where} \;\;
    \mathcal{KG}_{\text{global}} = \bigcup\limits_{i=1}^{t-1} \mathcal{KG}_{i}
\end{equation}
Here $\mathcal{KG}_{\textrm{global}}$ is the set of all edges that the agent has ever had in its knowledge graph and the subtraction operator is a set difference.
When the agent adds new edges to the graph perhaps as a the result of finding a new room $\mathcal{KG}_{\text{global}}$ changes and a positive reward is generated---this does not happen when that room is rediscovered in subsequent episodes.
This is then scaled by the game score so the intrinsic motivation does not drown out the actual quest rewards, the overall reward the agent receives at time step $t$ looks like this:
\begin{equation}
    r_t = r_{g_t} + \alpha r_{\text{IM}_t} \frac{r_{g_t} + \epsilon}{r_{\text{max}}}
    \label{eq:rewscalc}
\end{equation}
\begin{wraptable}[28]{r}{0.6\textwidth}
\vspace{-2\baselineskip}
\begin{minipage}{0.6\textwidth}
      \begin{algorithm}[H]
      \footnotesize
        \caption{Structured Exploration}
        \label{alg:backtrack}
        \begin{algorithmic}
        \State $\{\pi_{\text{chain}}, \pi_b, \pi\} \gets \phi$ \Comment{Chained, backtrack, current policy}
        \State $\{\mathcal{S}_{b}, \mathcal{S}\} \gets \phi$ \Comment{Backtrack, current state buffers}
        \State $s_0, r_\textrm{init} \gets \textproc{Env.Reset()}$
        \State $\mathcal{J}_{\text{max}} \gets r_\textrm{init}, p \gets 0$ 
         \For{\textrm{timestep} t in 0...M} \Comment{Train for M Steps}
            \State $s_{t+1}, r_{t}, \pi \gets \textproc{\qbert{}update}(s_{t}, \pi)$
            \State $\mathcal{S} \gets \mathcal{S} + s_{t+1}$ \Comment{Append current state to state buffer}
            \State $p \gets p + 1$ \Comment{Lose patience}
            \If{$\mathcal{J}(\pi) \leq \mathcal{J}_{\text{max}}$}
                \If{$p >= patience$} \Comment{Stuck at a bottleneck}
                    \State $s_t, r_{\text{max}}, \pi \gets \textproc{Backtrack}(\pi_b, \mathcal{S}_{b})$
                    \State \Comment{Bottleneck passed; Add $\pi$ to the chained policy}
                    \State $\pi_{\text{chain}} \gets \pi_{\text{chain}} + \pi$
                \EndIf
            \EndIf
            \If{$\mathcal{J}(\pi) > \mathcal{J}_{\text{max}}$} \Comment{New highscore found}
                \State $\mathcal{J}_{\text{max}} \gets \mathcal{J}(\pi);\pi_b \gets \pi; \mathcal{S}_{b} \gets \mathcal{S}; p \gets 0$
            \EndIf
        \EndFor
        \Return $\pi_{\text{chain}}$ \Comment{Chained policy that reaches max score}
        \State
        \Function{Q*BERTupdate}{$s_{t}, \pi$} \Comment{One-step update}
                
            \State $s_{t+1},r_{g_{t}} \gets \textproc{Env.Step}(s_{t},\pi)$ \Comment{Section~\ref{sec:qbertall}}
            \State $r_{t} \gets \textproc{CalculateReward}(s_{t+1},r_{g_{t}})$  \Comment{Eq.~\ref{eq:rewscalc}}
            \State $\pi \gets \textproc{A2C.update}(\pi, r_{t})$ \Comment{Appendix~\ref{app:qbert}}
            \State \Return $s_{t+1}, r_{t}, \pi$
                
        \EndFunction
        \State
        \Function{Backtrack}{$\pi_b$, $\mathcal{S}_b$} \Comment{Try to overcome bottleneck}
        \For{$b$ in $\textproc{Reverse}(\mathcal{S}_b)$} \Comment{States leading to highscore}
            \State $s_0 \gets b; \pi \gets \phi$
            \For{timestep $t$ in 0...N} \Comment{Train for N steps}
                \State $s_{t+1}, r_{t}, \pi \gets \textproc{\qbert{}update}(s_{t}, \pi)$
                \If{$\mathcal{J}(\pi)>\mathcal{J}(\pi_b)$}
                    \Return $s_t, r_{t}, \pi$
                \EndIf
            \EndFor
        \EndFor
        \State \textbf{Terminate} \Comment{Can't find better score; Give up.}
        \EndFunction
        \end{algorithmic}
      \end{algorithm}
    \end{minipage}
\end{wraptable}
where $\epsilon$ is a small smoothing factor, $\alpha$ is a scaling factor, $r_{g_t}$ is the game reward, $r_{\text{max}}$ is the maximum score possible for that game, and $r_t$ is the reward received by the agent on time step $t$.
%
\subsection{Modular Policy Chaining} 


A primary reason that agents fail to pass bottlenecks is not  satisfying all the required dependencies.
To solve this problem, we introduce a method of policy chaining, where the agent utilizes the determinism of the simulator to backtrack to previously visited states in order to fulfill dependencies required to overcome a bottleneck.



Specifically, Algorithm~\ref{alg:backtrack} optimizes the policy $\pi$ as usual, but also keeps track of a buffer $\mathcal{S}$ of the distinct states and knowledge graphs that led up to each state
(we use state $s_t$ to colloquially refer to the combination of an observation $o_t$ and knowledge graph $\mathcal{K}\mathcal{G}_t$).
Similarly, a bottleneck buffer $\mathcal{S}_{b}$ and policy $\pi_b$ reflect the sequence of states and policy with the maximal return $\mathcal{J}_\textrm{max}$.
A bottleneck is identified when the agents fails to improve upon $\mathcal{J}_\textrm{max}$ after \textit{patience} number of steps, i.e. no improvement in raw game score or knowledge-graph-based intrinsic motivation reward.
The agent then {\em backtracks} by searching backwards through the state sequence $\mathcal{S}_b$, restarting from each of the previous states---and training for $N$ steps in search of a more optimal policy to overcome the bottleneck.
When such a policy is found, it is appended to modular policy chain $\pi_\textrm{chain}$.
Conversely, if no such policy is found, then we have failed to pass the current bottleneck and the training terminates.

\begin{table}[]
\centering
\scriptsize
\begin{tabular}{l|r|r||r|r|r|r||r|r|r}
\textbf{Expt.}    &  \multicolumn{2}{c||}{\textbf{Jericho-QA}}      & \multicolumn{2}{c|}{\textbf{KG-A2C}} & \multicolumn{2}{c||}{\textbf{Q*BERT}} & \multicolumn{2}{c|}{\textbf{MC!Q*BERT}}  & \textbf{GO!Q*BERT} \\ \hline
\textbf{Game Reward}        & \multicolumn{2}{c||}{} & \multicolumn{2}{c|}{\checkmark}        & \multicolumn{2}{c||}{\checkmark}        & \checkmark           & \checkmark           & \checkmark           \\
\textbf{Intrinsic Motive} & \multicolumn{2}{c||}{}     & \multicolumn{2}{c|}{}                             & \multicolumn{2}{c||}{}                                 &                                     & \checkmark           & \checkmark           \\ \hline
\textbf{Metric}               & \textbf{EM} & \textbf{F1}                & \textbf{Eps.} & \textbf{Max} & \textbf{Eps.} & \textbf{Max} & \textbf{Max}                       & \textbf{Max}                       & \textbf{Max}                       \\ \hline
zork1                            &              40.01                &            44.62              & 34                &               35                    & 33.6              &              35                    &                         32            &                    \textbf{41.6}                 &        31                            \\
library                           &               36.76               &              46.45             & 14.3              &             19                     & 10.0              &                  18                 &                           19          &                   19                  &           18                          \\
detective                &               60.28               &                63.21                      & 207.9             &               214                    & 246.1             &              274                     &                      320               &                   \textbf{330}                  &           304                          \\
balances                       &               55.26               &                56.49                     & 10                &                 10                  & 9.8               &                  10                 &                        10             &                     10                &             10                        \\
pentari                        &               63.89               &               68.37                & 50.7              &                 56                  & 48.2              &               56                    &                            56         &                   \textbf{58}                 &              40                       \\
ztuu                        &              28.71                &              29.76                      & 6                 &                   9                & 5                 &                     5              &                              5       &                  \textbf{11.8}                   &                 5                    \\
ludicorp                     &               52.32               &              59.95                     & 17.8              &              19                     & 17.6              &                19                   &                             19        &                  \textbf{22.8}                   &             20.6                        \\
deephome                  &                8.03              &               9.27                      & 1                 &                  1                 & 1                 &                 1                  &                              \textbf{8}       &                   6                  &               1                      \\
temple                   &              48.92                &                49.17                      & 7.6               &                8                   & 7.9               &                  8                 &                                8     &                   8                  &                8                    
\end{tabular}
\caption{QA results on Jericho-QA test set and averaged asymptotic scores on games by different methods across 5 independent runs. For KG-A2C and \qbert{}, we present scores averaged across the final 100 episodes as well as {\em max scores}. Methods using exploration strategies show only {\em max scores} given their workings. Agents are allowed $10^6$ steps for each parallel A2C agent with a batch size of 16.}
\label{tab:scores}
\end{table}

\section{Evaluation}
We first evaluate the quality of the knowledge graph construction in a supervised setting.
Next we perform and end-to-end evaluation in which knowledge graph construction is used by \qbert.
\subsection{Graph Extraction Evaluation}
\begin{figure*} 
    \centering
    \begin{subfigure}{0.49\textwidth}
    \centering
    \includegraphics[width=\linewidth]{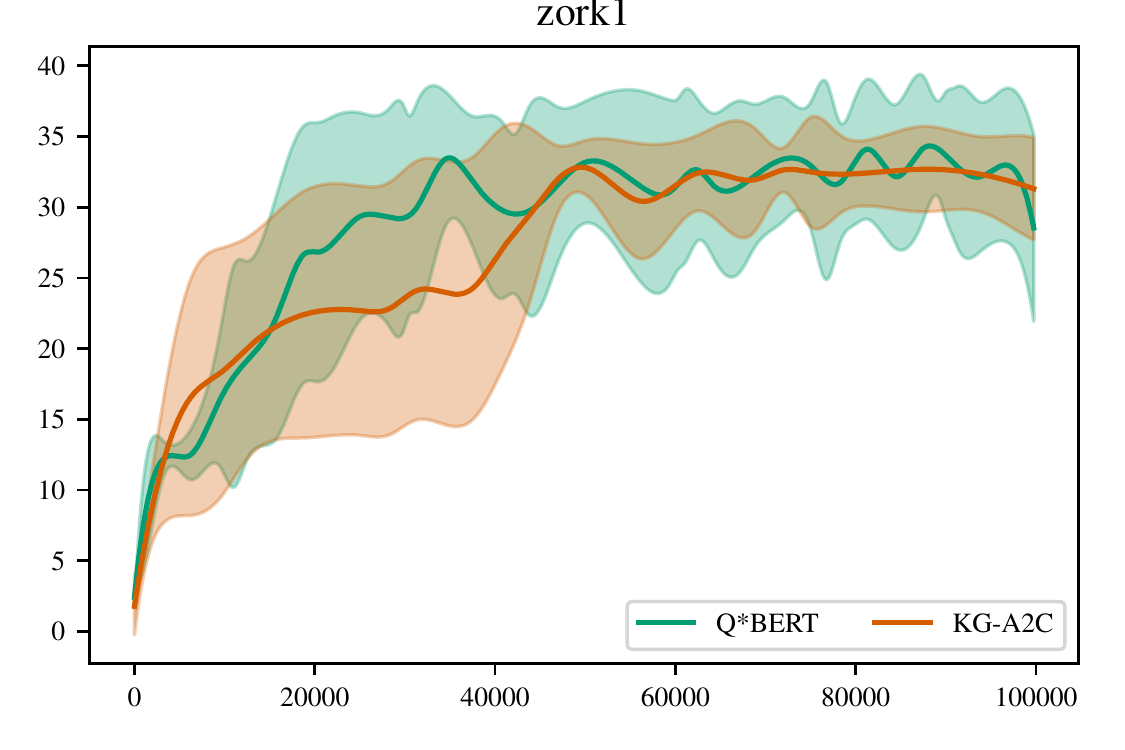}
    \caption{Episode reward curves for KG-A2C and Q*BERT.}
    \label{fig:graphcomp}
    \end{subfigure}
    \begin{subfigure}{0.49\textwidth}
    \centering
    \includegraphics[width=\linewidth]{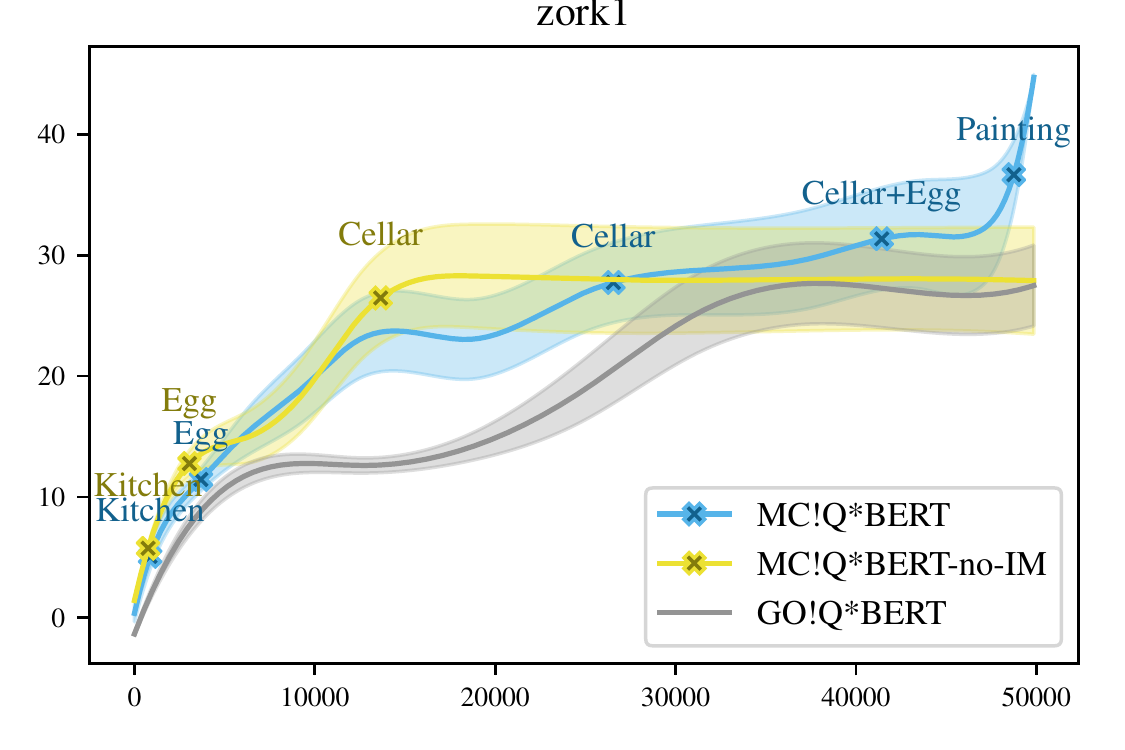}
    \caption{Max reward curves for exploration strategies.}
    \label{fig:explorecomp}
    \end{subfigure}
    \caption{Select ablation results on {\em Zork1} conducted across 5 independent runs per experiment. We see where the agents using structured exploration pass each bottleneck seen in Fig.~\ref{fig:dag}. \qbert{} without IM is unable to detect nor surpass bottlenecks beyond the {\em Cellar}.}
\end{figure*}

Table~\ref{tab:scores} left shows QA performance on the Jericho-QA dataset.
Exact match (EM) refers to the percentage of times the model was able to predict the exact answer string, while F1 measures token overlap between prediction and ground truth.
We observe
a direct correlation between the quality of the extracted graph and \qbert{}'s performance on the games.
On games where \qbert{} performed comparatively better than KG-A2C in terms of asymptotic scores, e.g. {\em detective}, the QA model had relatively high EM and F1, and vice versa as seen with {\em ztuu}.
In general, however, \qbert{} reaches comparable asymptotic performance to KG-A2C on 7 out of 9 games.
However, as shown in Figure~\ref{fig:graphcomp}, \qbert{} reaches asymptotic performance faster than KG-A2C, indicating that the QA model leads to faster learning.
Appendix~\ref{app:results} contains more plots illustrating this trend.
Both agents rely on the graph to constrain the action space and provide a richer input state representation.
\qbert{} uses a QA model fine-tuned on regularities of a text-game producing more relevant knowledge graphs than those extracted by OpenIE~\cite{Angeli2015} in KG-A2C for this purpose.

\subsection{Intrinsic Motivation and Exploration Strategy Evaluation}


We evaluate intrinsic motivation through policy chaining, dubbed \textbf{MC!Q*BERT} (Modularly Chained \qbert{}) by first testing policy chaining with only game reward and with both game reward and intrinsic motivation.
We provide a qualitative analysis of the bottlenecks detected with both methods with respect to those found in Fig.~\ref{fig:dag} on {\em Zork1}.
Just as KG-A2C provided us with a direct comparison for assessing the graph extraction abilities of \qbert{}, we test MC!\qbert{}'s structured exploration against an alternative exploration method in the form of Go-Explore~\citep{ecoffet19},
an algorithm with similar properties to our policy module chaining and has been show to work well for large, discrete game state spaces. 
Further, MC!\qbert{} using both game reward and intrinsic motivation matches or outperforms all other methods on 8 out of 9 games, with MC!\qbert{} using only game reward received the highest score on the $9^{th}$ game, {\em deephome}.

\textbf{GO!Q*BERT}
Go-Explore~\citep{ecoffet19} is an algorithm designed to keep track of sub-optimal and under-explored states in order to allow the agent to explore upon more optimal states that may be a result of sparse rewards.
The Go-Explore algorithm consists of two phases, the first to continuously explore until a set of promising states and corresponding trajectories are found on the basis of total score, and the second to robustify this found policy against potential stochasticity in the game.
Promising states are defined as those states when explored from will likely result in higher reward trajectories. 
\citet{madotto2020exploration} look at applying Go-Explore to text-games on a set of simpler games generated using the game generation framework TextWorld~\citep{cote18}.
They use a small set of ``admissible actions''---actions guaranteed to change the world state at any given step during Phase 1---to explore and find high reward trajectories.
We adapt this, instead training \qbert{} in parallel to generate actions from the full action space used for exploration to maintain a constant action space size across all models.
Implementation details are found in Appendix~\ref{app:goqbert}.

\section{Analysis}
Table~\ref{tab:scores} shows that across all the games MC!\qbert{} matches or outperforms the current state-of-the-art 
when compared across the metric of the max score consistently received across runs.
There are two main trends:
First, MC!\qbert{} greatly benefits from the inclusion of intrinsic motivation rewards.
Qualitative analysis of bottlenecks detected by each agent on the game of {\em Zork1} reveals differences in the overall accuracy of the bottleneck detection between MC!\qbert{} with and without intrinsic motivation.
Figure~\ref{fig:explorecomp} shows exactly when each of these agents detects and subsequently overcomes the bottlenecks outlined in Figure~\ref{fig:dag}.
What we see here is that when the intrinsic motivation is not used, the agent discovers that it can get to the {\em Kitchen} with a score of $+10$ and then {\em Cellar} with a score of $+25$ immediately after.
It forgets how to get the {\em Egg} with a smaller score of $+5$ and never makes it past the Grue in the {\em Cellar}.
Intrinsic motivation prevents this in two ways: (1)~it makes it less focused on a locally high-reward trajectory---making it less greedy and helping it chain together rewards for the {\em Egg} and {\em Cellar}, and (2)~provides rewards for fulfilling dependencies that would otherwise not be rewarded by the game---this is seen by the fact that it learns that picking up the lamp is the right way to surpass the {\em Cellar} bottleneck and reach the {\em Painting}.
A similar behavior is observed with GO!\qbert{}, the agent settles pre-maturely on a locally high-reward trajectory and thus never has incentive to find more globally optimal trajectories by fulfilling the underlying dependency graph.
Here, the likely cause is due to GO!\qbert{}'s inability to backtrack and rethink discovered high reward trajectories.

The second point is that using both the improvements to graph construction in addition to intrinsic motivation and structured exploration consistently yields higher max scores across a majority of the games when compared to the rest of the methods.
Having just the improvements to graph building or structured exploration by themselves is not enough.
Thus we infer that the full MC!\qbert{} agent is fundamentally exploring this combinatorially-sized space more effectively by virtue of being able to more consistently detect and clear bottlenecks.
The improvement over systems using default exploration such as KG-A2C or \qbert{} by itself indicates that structured exploration is necessary when dealing with sparse and ill-placed reward functions.


\section{Conclusions}

Modern deep reinforcement learning agents using default exploration strategies such as $\epsilon$-greedy are ill-equipped to deal with the challenge of sparse and delayed rewards, especially when placed in a combinatorially-sized state-action space.
Building on top of \qbert{}, an agent that constructs a knowledge graph of the world by asking questions about it, we introduce MC!\qbert{}, an agent that uses this graph as an intrinsic motivation to help detect bottlenecks arising from delayed rewards and chains policies that go from bottleneck to bottleneck.
A key insight from an ablation study is that the graph-based intrinsic motivation is crucial for bottleneck detection, preventing the agent from falling into locally optimal high reward trajectories due to ill-placed rewards.
Policy chaining used in tandem with intrinsic motivation results in agents that explore further in the game by clearing bottlenecks more consistently.

\section{Broader Impacts}

The ability to plan for long-term state dependencies in partially-observable environment has downstream applications beyond playing games.
We see text games as simplified analogues for systems capable of long-term dialogue with humans, such as in assistance with planning complex tasks, and also discrete planning domains such as logistics.
Broadly speaking, reinforcement learning is applicable to many sequential tasks, some of which cannot be anticipated.
Reinforcement learning for text environments are more suited for domains in which change in the world is affected via language, which mitigates physical risks---our line of work is not directly relevant to robotics---but not cognitive and emotional risks, as any system capable of generating natural language is capable of accidental or intentional non-normative language use~\cite{frazier:aies2019}.

\bibliography{neurips_2020.bib}
\bibliographystyle{abbrvnat}

\newpage
\appendix

\section{Implementation Details}
\label{appendix}

We would like to preface the appendix with a discussion on the relative differences in the assumptions that \qbert{} and MC!\qbert{} make regarding the underlying environment.
Although both are framed as POMDPs, MC!\qbert{} makes stronger assumptions regarding the determinism of the game as compared to \qbert{}.
MC!\qbert{} (and GO!\qbert{}) rely on the fact that the set of transition probabilities in a text-game are mostly deterministic.
Using this, they are able to assume that frozen policies can be executed deterministically, i.e. with no significant deviations from the original trajectory.
It is possible to robustify such policies by extending our method of structured exploration to perhaps perform imitation learning on the found highest score trajectories as seen in Phase 2 of the original GoExplore algorithm~\cite{ecoffet19}.
Stochasticity is not among set of challenges tackled in this work, however---we focus on learning how to better explore combinatorially-sized spaces with underlying long-term dependencies.
For future works in this space, we believe that agents should be compared based on the set of assumptions made: agents like KG-A2C and \qbert{} when operating under standard reinforcement learning assumptions, and MC!\qbert{} and GO!\qbert{} when under the stronger assumption of having a deterministic simulator.

\subsection{\qbert{}}
This section outlines how \qbert{} is trained, including details of the Jericho-QA dataset, the overall architecture, A2C training and hyperparameter details.
\label{app:qbert}
\subsubsection{Jericho-QA Dataset}
Jericho-QA contains 221453 Question-Answer pairs in the training set and 56667 pairs in the held out test set.
The test set consists of all the games that we test on in this paper.
It is collected by randomly exploring games using a set of admissible actions in addition to using the walkthroughs for each game as found in the Jericho framework~\citep{jericho}.
The set of attributes for a game is taken directly from the game engine and is defined by the game developer.

A single sample looks like this:
\begin{lstlisting}
Context: 
[loc] Chief's Office    You are standing in the chief's office. He is telling you, "The mayor was murdered yeaterday night at 12:03 am. I want you to solve it before we get any bad publicity or the FBI has to come in." "Yessir!" you reply. He hands you a sheet of paper. once you have read it, go north or west.  You can see a piece of white paper here.  
[inv]  You are carrying nothing.  
[obs]  [your score has just gone up by ten points.]  
[atr] talkable, seen, lieable, enterable, nodwarf, indoors, visited, handed, lockable, surface, thing, water_room, unlock, lost, afflicted, is_treasure, converse, mentioned, male, npcworn, no_article, relevant, scored, queryable, town, pluggable, happy, is_followable, legible, multitude, burning, room, clothing, underneath, ward_area , little, intact, animate, bled_in, supporter, readable, openable, near, nonlocal, door, plugged, sittable, toolbit, vehicle, light, lens_searchable, open, familiar, is_scroll, aimable, takeable, static, unique, concealed, vowelstart, alcoholic, bodypart, general, is_spell, full, dry_land, pushable, known, proper, inside, clean, ambiguously_plural, container, edible, treasure, can_plug, weapon, is_arrow, insubstantial, pluralname, transparent, is_coin, air_room, scenery, on, is_spell_book, burnt, burnable, auto_searched, locked, switchable, absent, rockable, beenunlocked, progressing, severed,     worn, windy, stone, random, neuter, legible, female, asleep, wiped
Question: Where am I located? Answer: chief's office
Question: What is here? Answer: paper, west
Question: What do I have? Answer: nothing
Question: What attributes does paper have? Answer: legible, animate
Question: What attributes does west have? Answer: room, animate
\end{lstlisting}

\newpage
\subsubsection{Knowledge Graph Update Rules}
\label{app:kgupdates}
Every step, given the current state and possible attributes as context---the QA network predicts the current room location, the set of all inventory objects, the set of all surrounding objects, and all attributes for each object.
\begin{itemize}
    \item Linking the current room type (e.g. ``Kitchen'', ``Cellar'') to the items found in the room with the relation ``has'',
        e.g. $\langle kitchen, has, lamp\rangle$
    \item All attribute information for each object is linked to the object with the relation ``is''.
        e.g. $\langle egg, is, treasure\rangle$
    \item Linking all inventory objects with relation ``have'' to the ``you'' node,
        e.g. $\langle you, have, sword\rangle$
    \item Linking rooms with directions based on the action taken to move between the rooms,
        e.g. $\langle Behind$ $House,east$ $of, Forest\rangle$ after the action ``go east'' is taken to go from behind the house to the forest
\end{itemize}

\begin{figure}
    \centering
    \includegraphics[width=.8\linewidth]{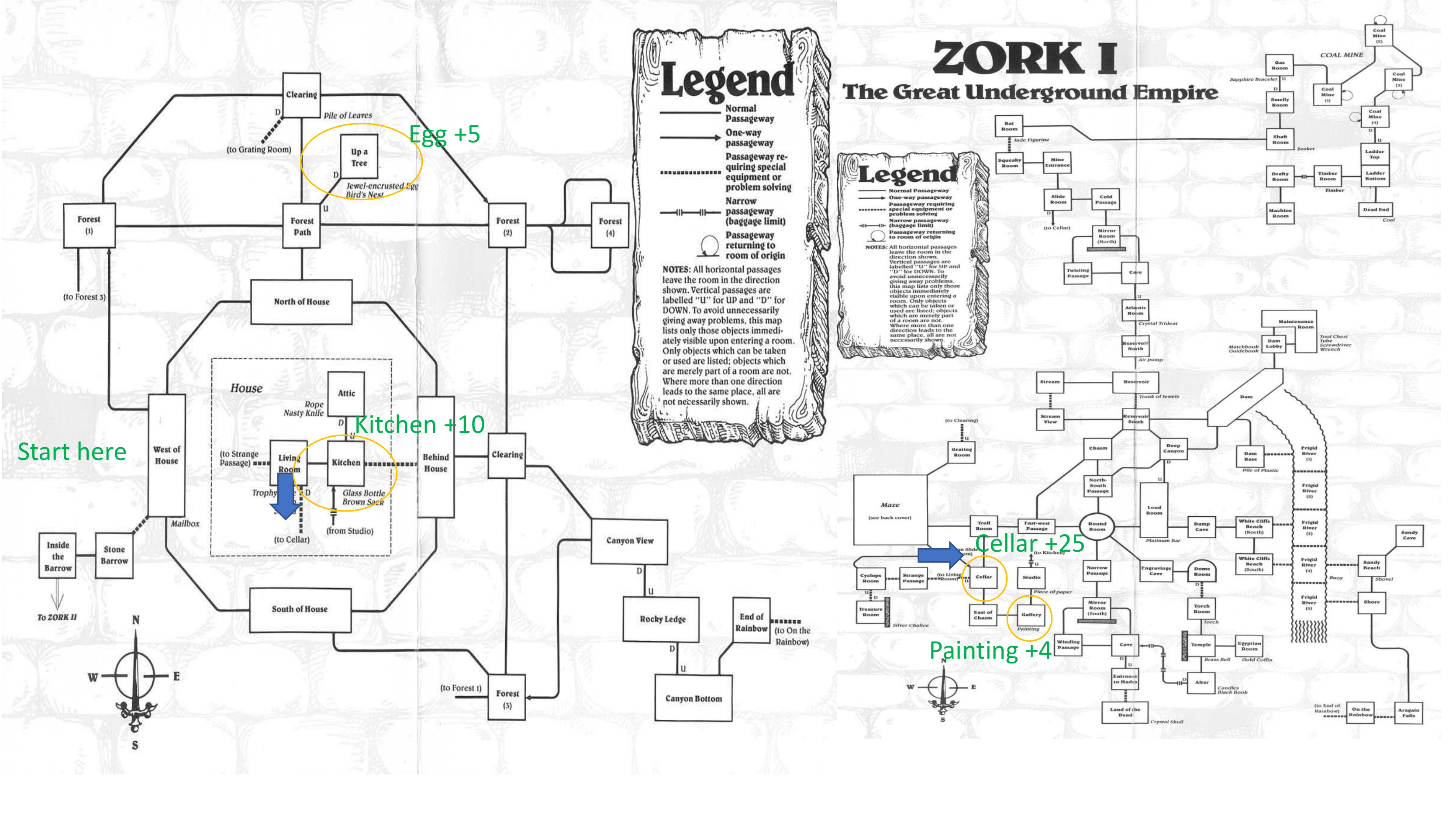}
    \caption{Map of {\em Zork1} annotated with rewards taken from \citet{Ammanabrolu2020Graph} and corresponding to the states and rewards found in Figure~\ref{fig:dag}.}
    \label{fig:zorkmap}
\end{figure}

Below is an excerpt from {\em Zork1} showing the exact observations given to the \qbert{},the knowledge graph, and the corresponding action taken by the agent after the graph extraction and update process has occurred as described above for a trajectory consisting of 5 timesteps.
These timesteps begin at the start of the game in {\em West of House} and continue till the agent has entered the {\em Kitchen} as seen in Fig.~\ref{fig:dag} and Fig.~\ref{fig:zorkmap}.
The set of $\langle s, r, o \rangle$ triples that make up the graph are in the text and the figure shows a partial visualization of the graph at that particular step in the trajectory.

\begin{minipage}{\textwidth}
  \includegraphics[width=0.6\textwidth, center]{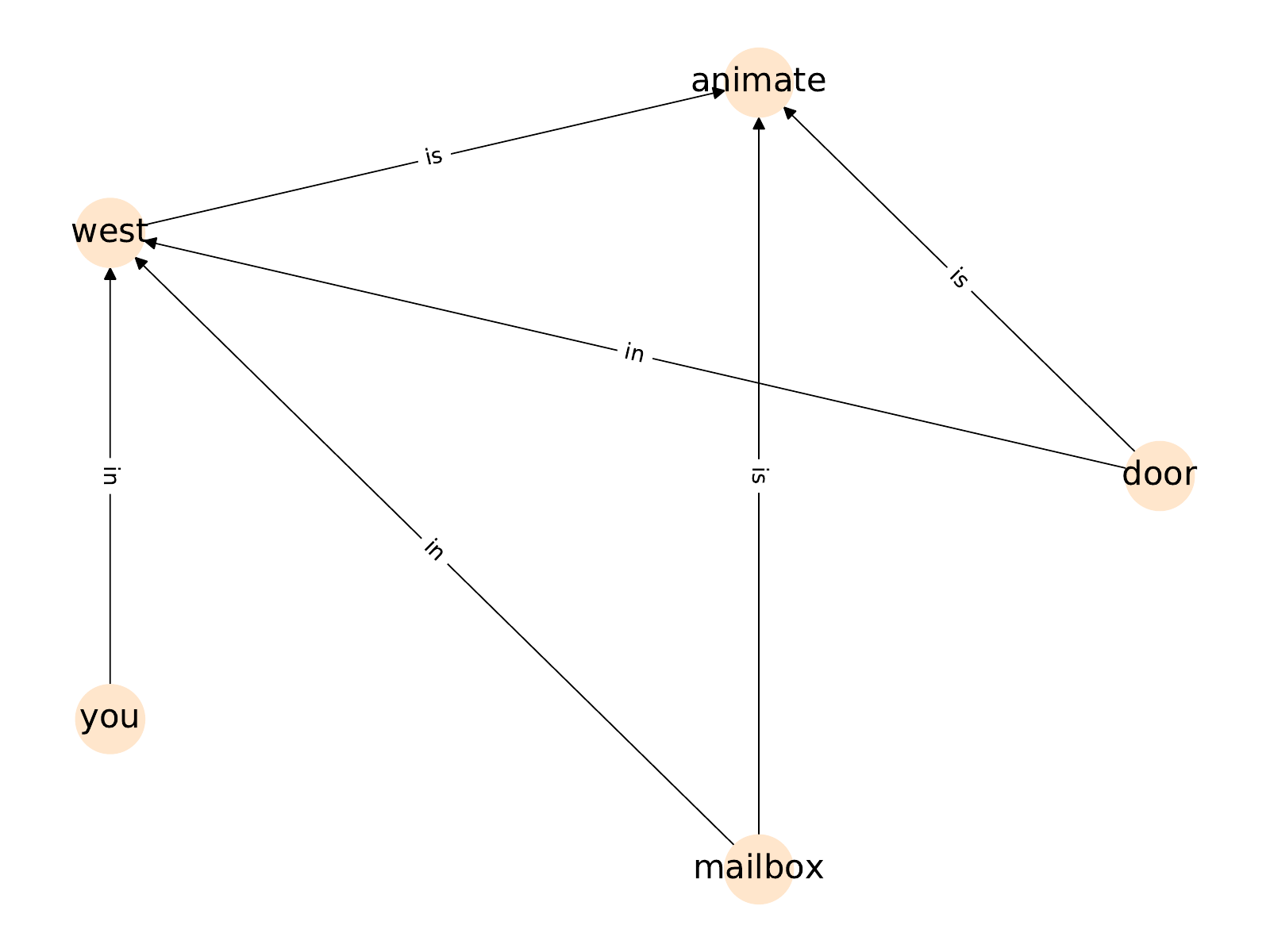}
    \begin{lstlisting}
[loc] West of House You are standing in an open field west of a white house, with a boarded front door. There is a small mailbox here.   [inv] You are empty handed.   
[obs] Copyright  c  1981, 1982, 1983 Infocom, Inc. All rights reserved. ZORK is a registered trademark of Infocom, Inc. Revision 88 / Serial number 840726  West of House You are standing in an open field west of a white house, with a boarded front door. There is a small mailbox here. 
[atr] talkable, seen, lieable, enterable, nodwarf, indoors, visited, handed, lockable, surface, thing, water_room, unlock, lost, afflicted, is_treasure, converse, mentioned, male, npcworn, no_article, relevant, scored, queryable, town, pluggable, happy, is_followable, legible, multitude, burning, room, clothing, underneath, ward_area, little, intact, animate, bled_in, supporter, readable, openable, near, nonlocal, door, plugged, sittable, toolbit, vehicle, light, lens_searchable, open, familiar, is_scroll, aimable, takeable, static, unique, concealed, vowelstart, alcoholic, bodypart, general, is_spell, full, dry_land, pushable, known, proper, inside, clean, ambiguously_plural, container, edible, treasure, can_plug, weapon, is_arrow, insubstantial, pluralname, transparent, is_coin, air_room, scenery, on, is_spell_book, burnt, burnable, auto_searched, locked, switchable, absent, rockable, beenunlocked, progressing, severed, worn, windy, stone, random, neuter, legible, female, asleep, wiped
[graph] [(you, in, west), (door, is, animate), (door, in, west), (west, is, animate),  (mailbox, in, west), (mailbox, is, animate)]
[next act] go north
\end{lstlisting}
\end{minipage}

\noindent
\begin{minipage}{\textwidth}
  \includegraphics[width=0.6\textwidth, center]{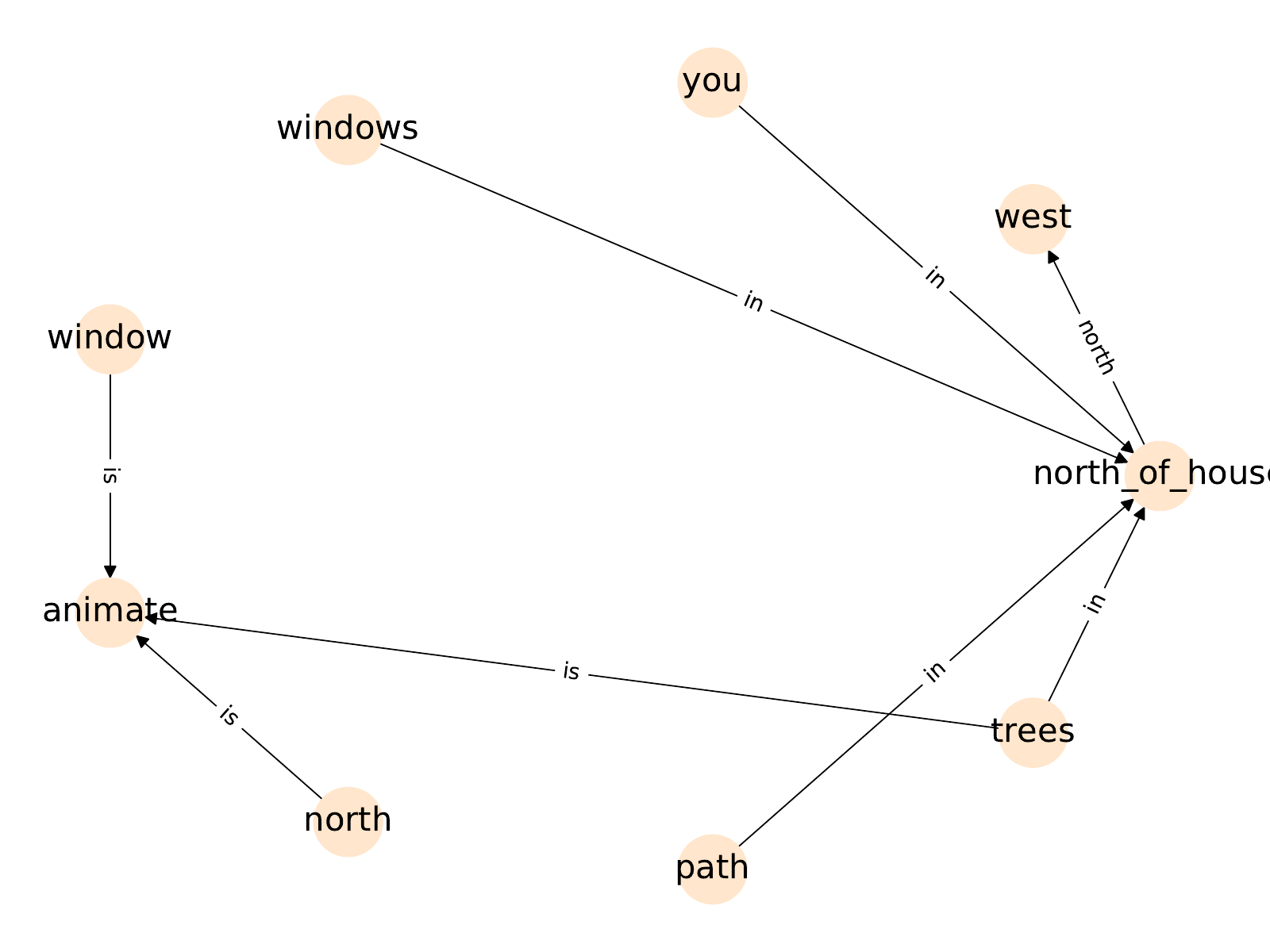}
    \begin{lstlisting}
[loc] North of House You are facing the north side of a white house. There is no door here, and all the windows are boarded up. To the north a narrow path winds through the trees.   
[inv] You are empty handed.   
[obs] North of House You are facing the north side of a white house. There is no door here, and all the windows are boarded up. To the north a narrow path winds through the trees. 
[atr] talkable, seen, lieable, enterable, nodwarf, indoors, visited, handed, lockable, surface, thing, water_room, unlock, lost, afflicted, is_treasure, converse, mentioned, male, npcworn, no_article, relevant, scored, queryable, town, pluggable, happy, is_followable, legible, multitude, burning, room, clothing, underneath, ward_area, little, intact, animate, bled_in, supporter, readable, openable, near, nonlocal, door, plugged, sittable, toolbit, vehicle, light, lens_searchable, open, familiar, is_scroll, aimable, takeable, static, unique, concealed, vowelstart, alcoholic, bodypart, general, is_spell, full, dry_land, pushable, known, proper, inside, clean, ambiguously_plural, container, edible, treasure, can_plug, weapon, is_arrow, insubstantial, pluralname, transparent, is_coin, air_room, scenery, on, is_spell_book, burnt, burnable, auto_searched, locked, switchable, absent, rockable, beenunlocked, progressing, severed, worn, windy, stone, random, neuter, legible, female, asleep, wiped
[graph] [(north_of_house, north, west),  (you, in, north_of_house), (door, is, animate), (door, in, west), (west, is, animate), (west, in, west),  (mailbox, in, west), (mailbox, is, animate), (windows, in, north_of_house), (windows, is, animate), (north, is, animate), (north, in, north_of_house), (path, is, animate), (path, in, north_of_house), (trees, in, north_of_house), (trees, is, animate)]
[next act] go east
\end{lstlisting}
\end{minipage}

\noindent
\begin{minipage}{\textwidth}
  \includegraphics[width=0.5\textwidth, center]{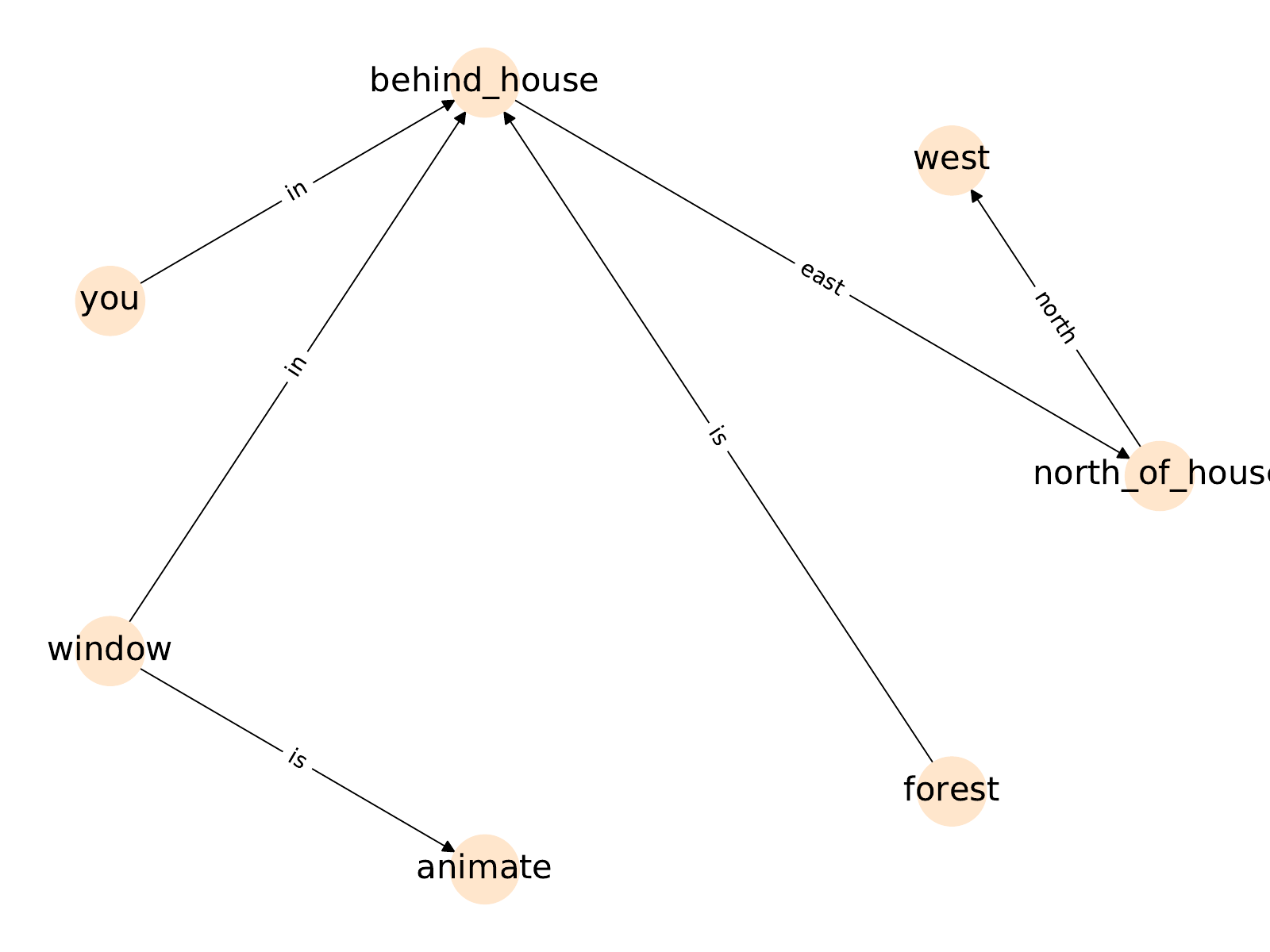}
    \begin{lstlisting}
[loc] Behind House You are behind the white house. A path leads into the forest to the east. In one corner of the house there is a small window which is slightly ajar.   
[inv] You are empty handed.   
[obs] Behind House You are behind the white house. A path leads into the forest to the east. In one corner of the house there is a small window which is slightly ajar. 
[atr] talkable, seen, lieable, enterable, nodwarf, indoors, visited, handed, lockable, surface, thing, water_room, unlock, lost, afflicted, is_treasure, converse, mentioned, male, npcworn, no_article, relevant, scored, queryable, town, pluggable, happy, is_followable, legible, multitude, burning, room, clothing, underneath, ward_area, little, intact, animate, bled_in, supporter, readable, openable, near, nonlocal, door, plugged, sittable, toolbit, vehicle, light, lens_searchable, open, familiar, is_scroll, aimable, takeable, static, unique, concealed, vowelstart, alcoholic, bodypart, general, is_spell, full, dry_land, pushable, known, proper, inside, clean, ambiguously_plural, container, edible, treasure, can_plug, weapon, is_arrow, insubstantial, pluralname, transparent, is_coin, air_room, scenery, on, is_spell_book, burnt, burnable, auto_searched, locked, switchable, absent, rockable, beenunlocked, progressing, severed, worn, windy, stone, random, neuter, legible, female, asleep, wiped
[graph] [(north_of_house, north, west), (behind_house, east, north_of_house),  (you, in, behind_house), (door, is, animate), (door, in, west), (west, is, animate), (west, in, west),  (you, in, behind_house), (mailbox, in, west), (mailbox, is, animate), (windows, in, north_of_house), (windows, is, animate), (north, is, animate), (north, in, north_of_house), (path, is, animate), (path, in, north_of_house), (trees, in, north_of_house), (trees, is, animate), (window, in, behind_house), (window, is, animate), (forest, in, behind_house), (forest, is, animate), (east, in, behind_house), (east, is, animate)]
[next act] open window
\end{lstlisting}
\end{minipage}

\noindent
\begin{minipage}{\textwidth}
  \includegraphics[width=0.5\textwidth, center]{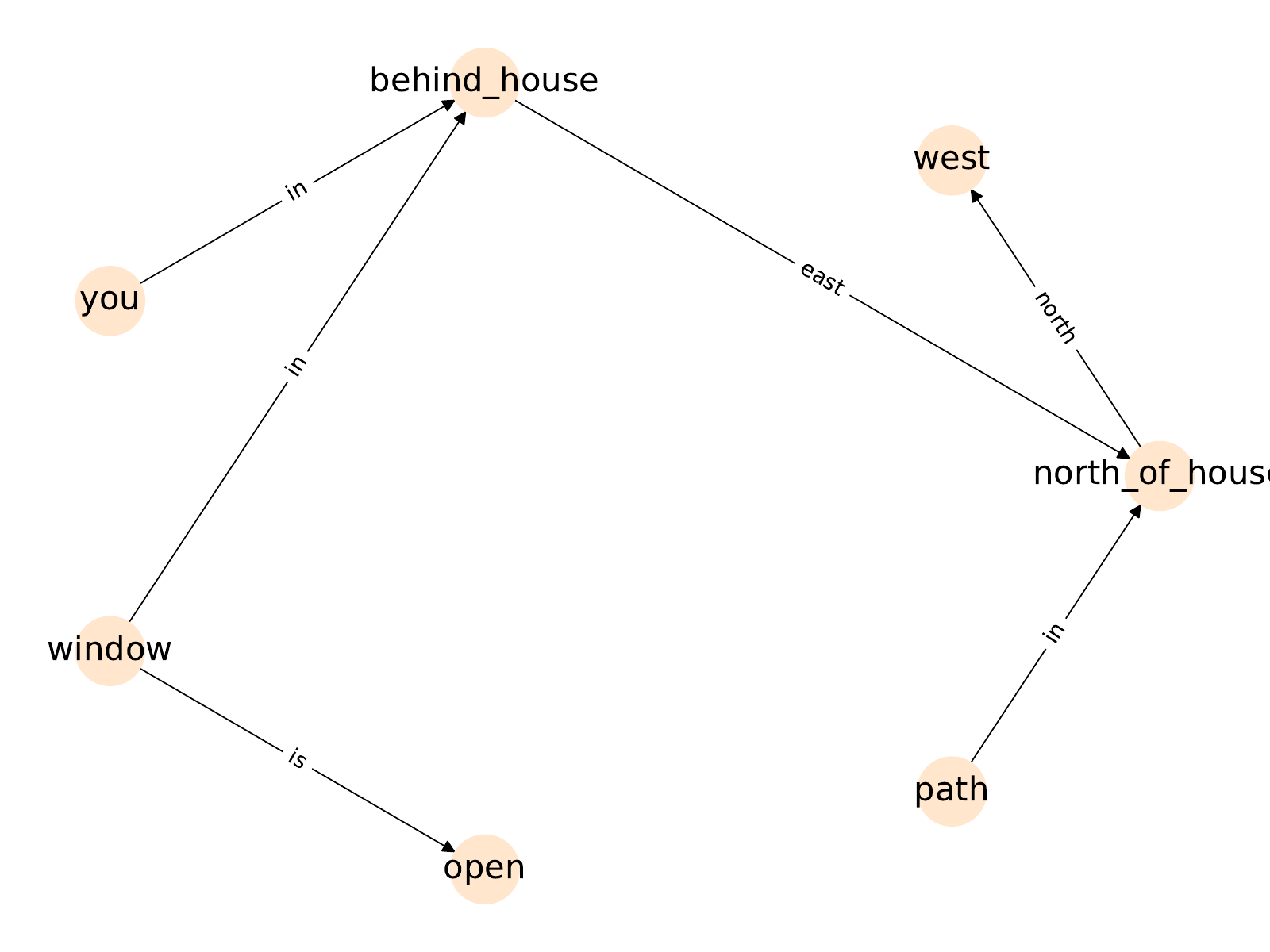}
    \begin{lstlisting}
[loc] Behind House You are behind the white house. A path leads into the forest to the east. In one corner of the house there is a small window which is open.   [inv] You are empty handed.   
[obs] With great effort, you open the window far enough to allow entry. 
[atr] talkable, seen, lieable, enterable, nodwarf, indoors, visited, handed, lockable, surface, thing, water_room, unlock, lost, afflicted, is_treasure, converse, mentioned, male, npcworn, no_article, relevant, scored, queryable, town, pluggable, happy, is_followable, legible, multitude, burning, room, clothing, underneath, ward_area, little, intact, animate, bled_in, supporter, readable, openable, near, nonlocal, door, plugged, sittable, toolbit, vehicle, light, lens_searchable, open, familiar, is_scroll, aimable, takeable, static, unique, concealed, vowelstart, alcoholic, bodypart, general, is_spell, full, dry_land, pushable, known, proper, inside, clean, ambiguously_plural, container, edible, treasure, can_plug, weapon, is_arrow, insubstantial, pluralname, transparent, is_coin, air_room, scenery, on, is_spell_book, burnt, burnable, auto_searched, locked, switchable, absent, rockable, beenunlocked, progressing, severed, worn, windy, stone, random, neuter, legible, female, asleep, wiped
[graph] [(north_of_house, north, west), (behind_house, east, north_of_house),  (you, in, behind_house), (door, is, animate), (door, in, west), (west, is, animate), (west, in, west),  (mailbox, in, west), (mailbox, is, animate), (windows, in, north_of_house), (windows, is, animate), (windows, is, open), (north, is, animate), (north, in, north_of_house), (path, is, animate), (path, in, north_of_house), (trees, in, north_of_house), (trees, is, animate), (window, in, behind_house), (window, is, animate), (forest, in, behind_house), (forest, is, animate), (east, in, behind_house), (east, is, animate)]
\end{lstlisting}
\end{minipage}

\noindent
\begin{minipage}{\textwidth}
  \includegraphics[width=0.5\textwidth, center]{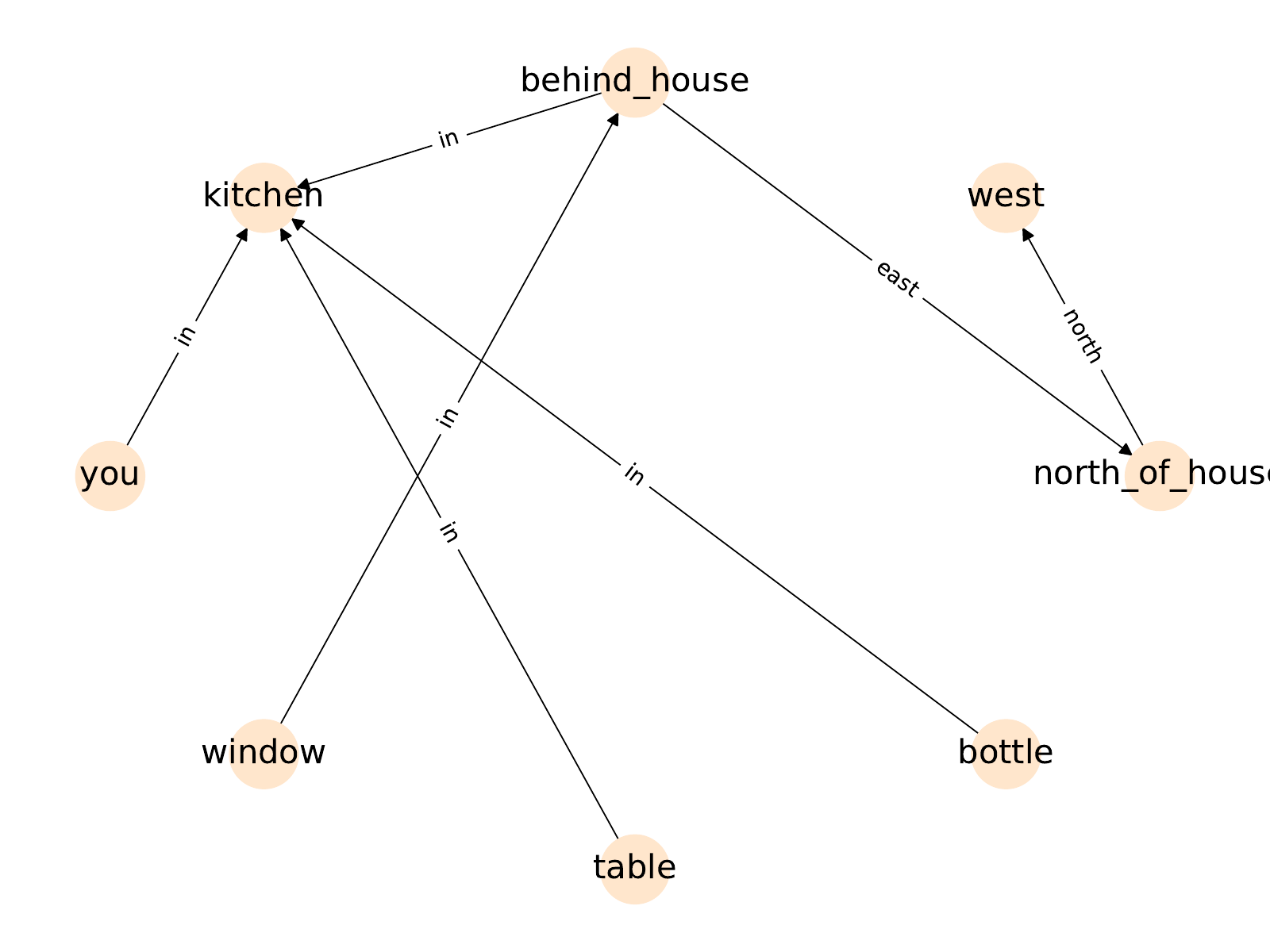}
    \begin{lstlisting}
[loc] Kitchen You are in the kitchen of the white house. A table seems to have been used recently for the preparation of food. A passage leads to the west and a dark staircase can be seen leading upward. A dark chimney leads down and to the east is a small window which is open. On the table is an elongated brown sack, smelling of hot peppers. A bottle is sitting on the table. The glass bottle contains:   A quantity of water   
[inv] You are empty handed.   
[obs] Kitchen You are in the kitchen of the white house. A table seems to have been used recently for the preparation of food. A passage leads to the west and a dark staircase can be seen leading upward. A dark chimney leads down and to the east is a small window which is open. On the table is an elongated brown sack, smelling of hot peppers. A bottle is sitting on the table. The glass bottle contains:   A quantity of water 
[atr] talkable, seen, lieable, enterable, nodwarf, indoors, visited, handed, lockable, surface, thing, water_room, unlock, lost, afflicted, is_treasure, converse, mentioned, male, npcworn, no_article, relevant, scored, queryable, town, pluggable, happy, is_followable, legible, multitude, burning, room, clothing, underneath, ward_area, little, intact, animate, bled_in, supporter, readable, openable, near, nonlocal, door, plugged, sittable, toolbit, vehicle, light, lens_searchable, open, familiar, is_scroll, aimable, takeable, static, unique, concealed, vowelstart, alcoholic, bodypart, general, is_spell, full, dry_land, pushable, known, proper, inside, clean, ambiguously_plural, container, edible, treasure, can_plug, weapon, is_arrow, insubstantial, pluralname, transparent, is_coin, air_room, scenery, on, is_spell_book, burnt, burnable, auto_searched, locked, switchable, absent, rockable, beenunlocked, progressing, severed, worn, windy, stone, random, neuter, legible, female, asleep, wiped
[graph] [(north_of_house, north, west), (behind_house, east, north_of_house), (behind_house, in, kitchen), (you, in, kitchen), (door, is, animate), (door, in, west), (west, is, animate), (west, in, west), (west, in, kitchen), (mailbox, in, west), (mailbox, is, animate), (windows, in, north_of_house), (windows, is, animate), (north, is, animate), (north, in, north_of_house), (path, is, animate), (path, in, north_of_house), (trees, in, north_of_house), (trees, is, animate), (window, in, behind_house), (window, is, animate), (forest, in, behind_house), (forest, is, animate), (east, in, behind_house), (east, is, animate), (table, in, kitchen), (table, is, animate)]]
[next act] go in
\end{lstlisting}
\end{minipage}

\subsubsection{Architecture}
Further details of what is found in Figure~\ref{fig:architecture}.
The sequential action decoder consists two GRUs that are linked together as seen in \citet{Ammanabrolu2020Graph}.
The first GRU decodes an action template and the second decodes objects that can be filled into the template.
These objects are constrained by a {\em graph mask}, i.e. the decoder is only allowed to select entities that are already present in the knowledge graph.

The question answering network based on ALBERT~\citep{Lan2020ALBERT:} has the following hyperparameters, taken from the original paper and known to work well on the SQuAD~\cite{rajpurkar-etal-2016-squad} dataset.
No further hyperparameter tuning was conducted.
\begin{center}
\begin{tabular}{l|r}
    \textbf{Parameters} & \textbf{Value}  \\ \hline 
     batch size & 8\\
     learning rate & \num{3e-5}\\
     max seq len & 512\\
     doc stride & 128\\
     warmup steps & 814\\
     max steps & 8144\\
     gradient accumulation steps & 24\\
 \end{tabular}
\label{tab:qbertParams}
\end{center}

\subsubsection{A2C Training}
The rest of the A2C training is unchanged from \citet{Ammanabrolu2020Graph}.
A2C training starts with calculating the advantage of taking an action in a state $A(s_t, a_t)$, defined as the value of taking an action $Q(s_t, a_t)$ compared to the average value of taking all possible {\em admissible actions} in that state $V(s_t)$:
\begin{align}
    A(s_t, a_t) = Q(s_t, a_t) - V(s_t)\\
    Q(s_t, a_t) = \mathbb{E}[r_t + \gamma V(s_{t+1})]
\end{align}
The value is predicted by the critic as shown in Fig.~\ref{fig:architecture} and $r_t$ is the reward received at step $t$.

The action decoder or actor is then updated according to the gradient:
\begin{equation}
    -\nabla_\theta(log \pi_\mathbb{T}(\tau|s_t;\theta_t) + \sum_{i=1}^{n}log\pi_{\mathbb{O}_i}(o_i|s_t,\tau,...,o_{i-1};\theta_t))A(s_t,a_t)
\end{equation}
updating the template policy $\pi_\mathbb{T}$ and object policies $\pi_{\mathbb{O}_i}$ based on the fact that each step in the action decoding process is conditioned on all the previously decoded portions.
The critic is updated with respect to the gradient:
\begin{equation}
    \frac{1}{2}\nabla_\theta(Q(s_t, a_t;\theta_t)-V(s_t;\theta_t))^2
\end{equation}
bringing the critic's prediction of the value of being in a state closer to its true underlying value.
An entropy loss is also added:
\begin{equation}
    \mathcal{L}_\mathbb{E}(s_t,a_t;\theta_t)=\sum_{a \in V(s_t)}P(a|s_t)logP(a|s_t)
\end{equation}
Hyperparameters are taken from KG-A2C as detailed by \citet{Ammanabrolu2020Graph} and not tuned any further.

\subsection{MC!\qbert{}}
The additional hyperparamters used for modular policy chaining are detailed below.
{\em Patience batch factor} is the proportion of the batch that must have stagnated at a particular score for {\em patience} number of episodes of unchanging score before a bottleneck is detected.
{\em Patience} within a range of $1000-6000$ in increments of $500$ and {\em buffer size} within a range of $10-60$ in increments of $10$ were the only additional parameters tuned for, on {\em Zork1}.
The resulting best hyperparameter set was used on the rest of the games.
\begin{center}
\begin{tabular}{l|c}
    \textbf{Parameters} & \textbf{Value}  \\ \hline 
     patience & 3000\\
     buffer size & 40\\
     batch size & 16\\
     patience batch factor & .75
 \end{tabular}
\label{tab:mcqbertParams}
\end{center}


\subsection{GO!\qbert{}}
%

\label{app:goqbert}
Since the text games we are dealing with are mostly deterministic, with the exception of {\em Zork1} in later stages, we only focus on using Phase 1 of the Go-Explore algorithm to find an optimal policy.
Go-Explore maintains an archive of cells---defined as a set of states that map to a single representation---to keep track of promising states.
\citet{ecoffet19} simply encodes each cell by keeping track of the agent's position and \citet{madotto2020exploration} use the textual observations encoded by recurrent neural network as a cell representation.
We improve on this implementation by training the \qbert{} network in parallel, using the snapshot of the knowledge graph in conjunction with the game state to further encode the current state and use this as a cell representation.
At each step, Go-Explore chooses a cell to explore at random (weighted by score to prefer more advanced cells).
\qbert{} will run for a number of steps in each cell, for all our experiments we use a cell step size of 32, starting with the knowledge graph state and the last seen state of the game from the cell.
This will generate a trajectory for the agent while further training \qbert{} at each iteration, creating a new representation for the knowledge graph as well as a new game state for the cell.
After expanding a cell, Go-Explore will continue to sample cells by weight to continue expanding its known states.
At the same time, \qbert{} will benefit from the heuristics of selecting preferred cells and be trained on promising states more often.

\section{Results}
\label{app:results}
\subsection{Graph Evaluation Results}
\begin{figure}[h]
\begin{minipage}{.33\textwidth}
  \centering
\includegraphics[width=\linewidth]{figures/qbert_base_gameonly/zork1_episode_score.pdf}\end{minipage}
\begin{minipage}{.33\textwidth}
  \centering
  \includegraphics[width=\linewidth]{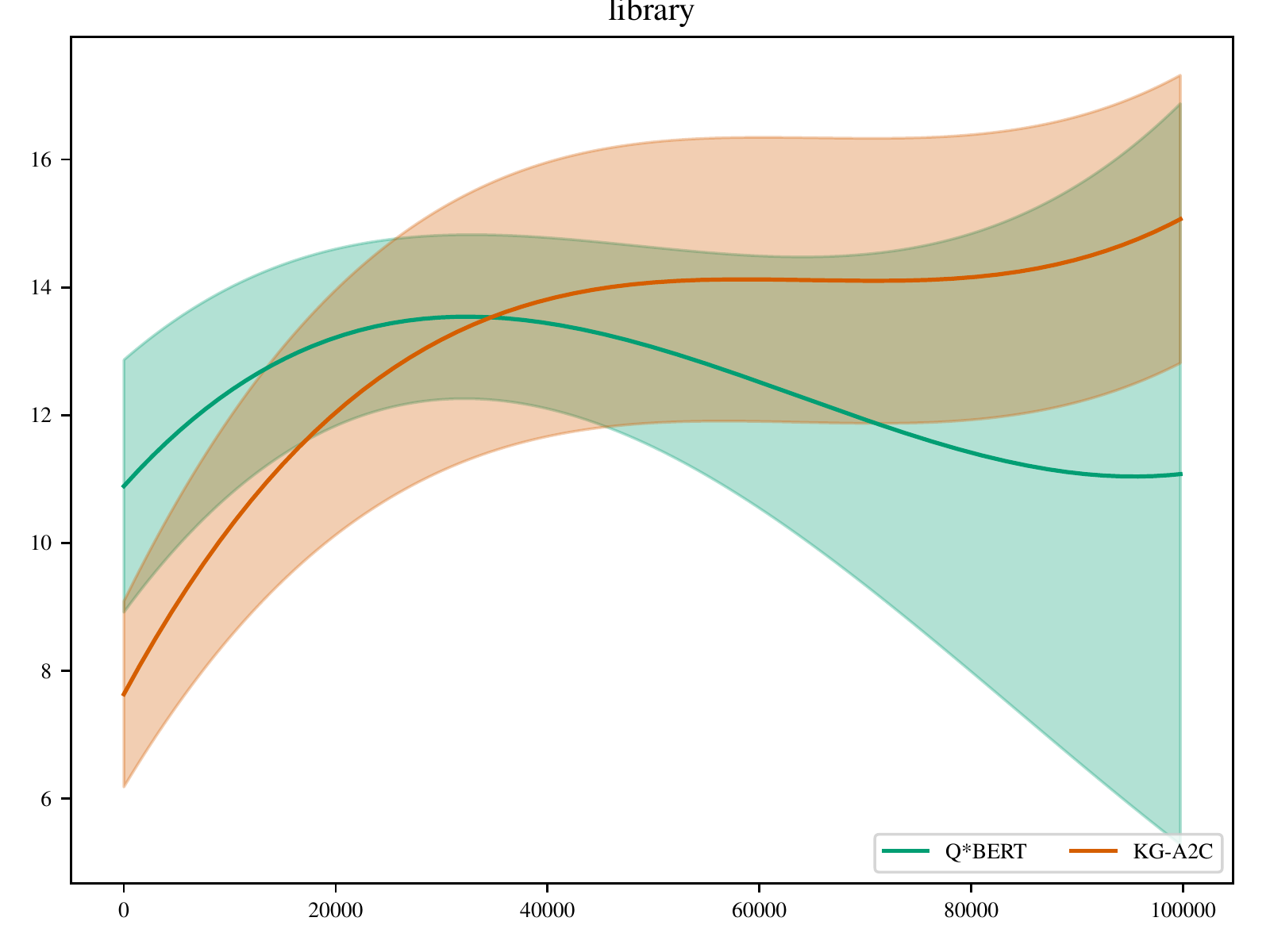}
\end{minipage}
\begin{minipage}{.33\textwidth}
  \centering
  \includegraphics[width=\linewidth]{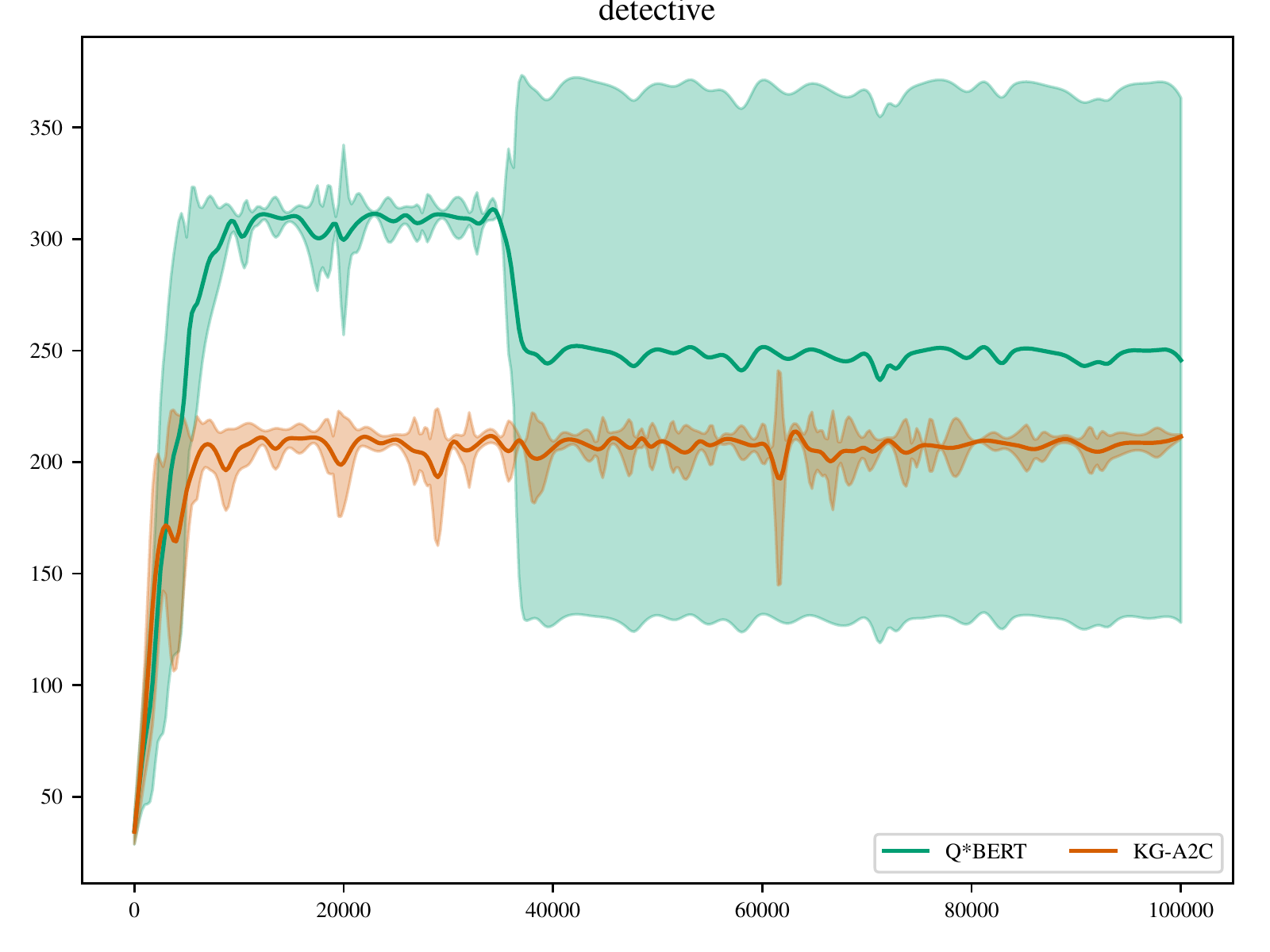}
\end{minipage}
\begin{minipage}{.33\textwidth}
  \centering
  \includegraphics[width=\linewidth]{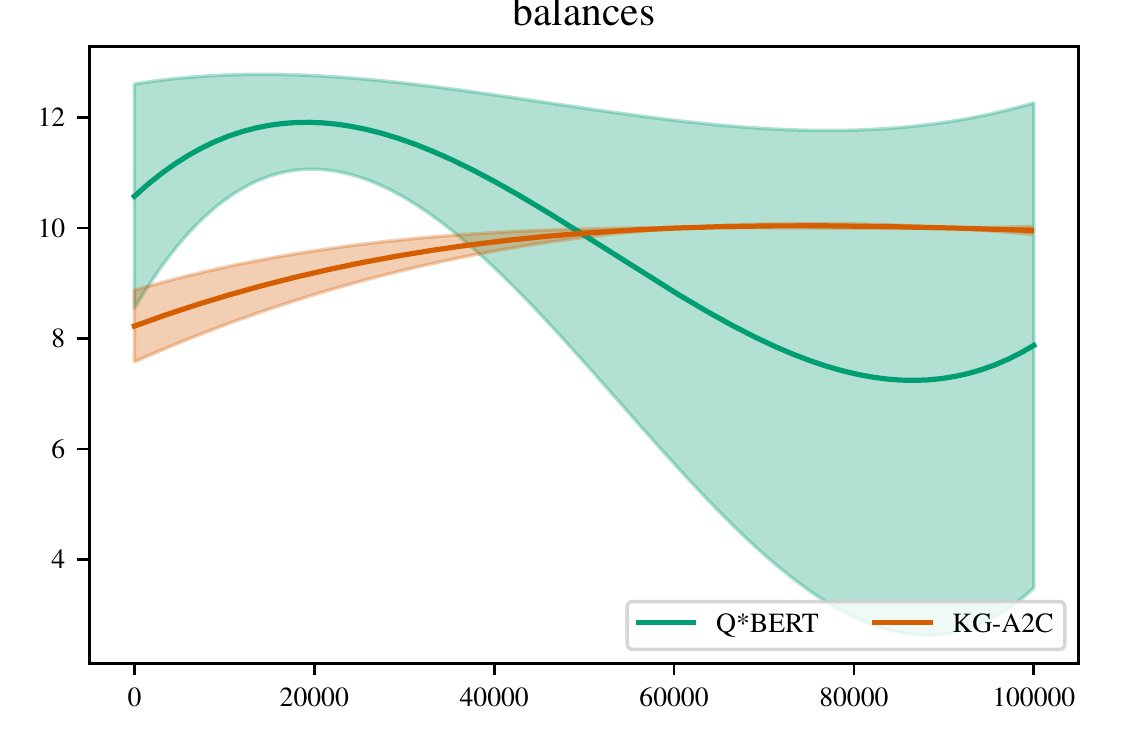}
\end{minipage}
\begin{minipage}{.33\textwidth}
  \centering
  \includegraphics[width=\linewidth]{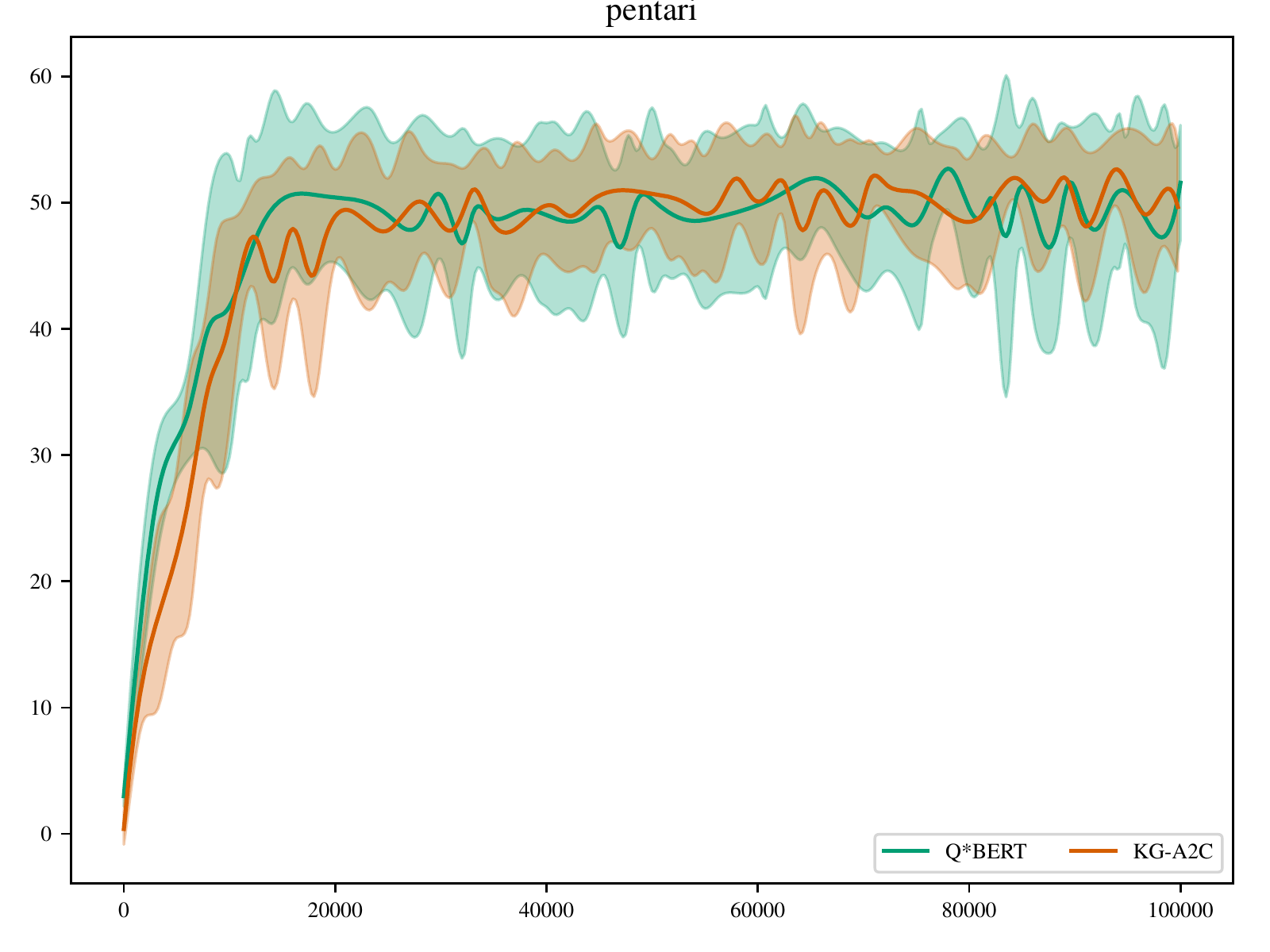}
\end{minipage}
\begin{minipage}{.33\textwidth}
  \centering
  \includegraphics[width=\linewidth]{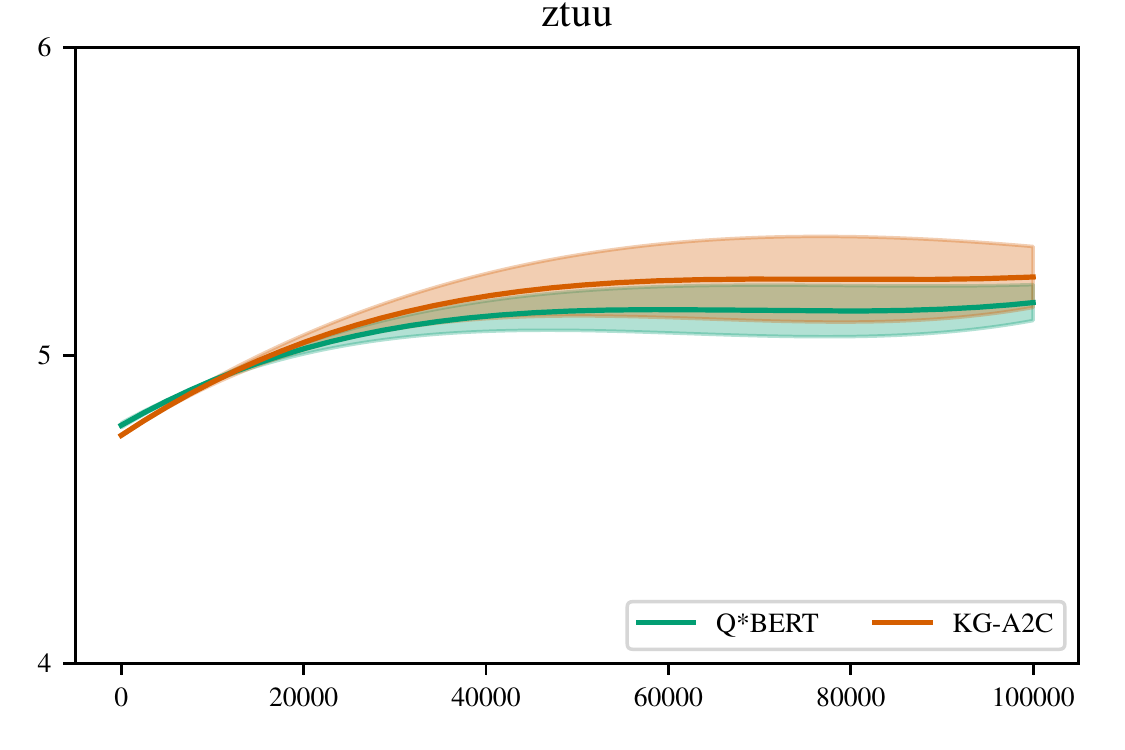}
\end{minipage}
\begin{minipage}{.33\textwidth}
  \centering
  \includegraphics[width=\linewidth]{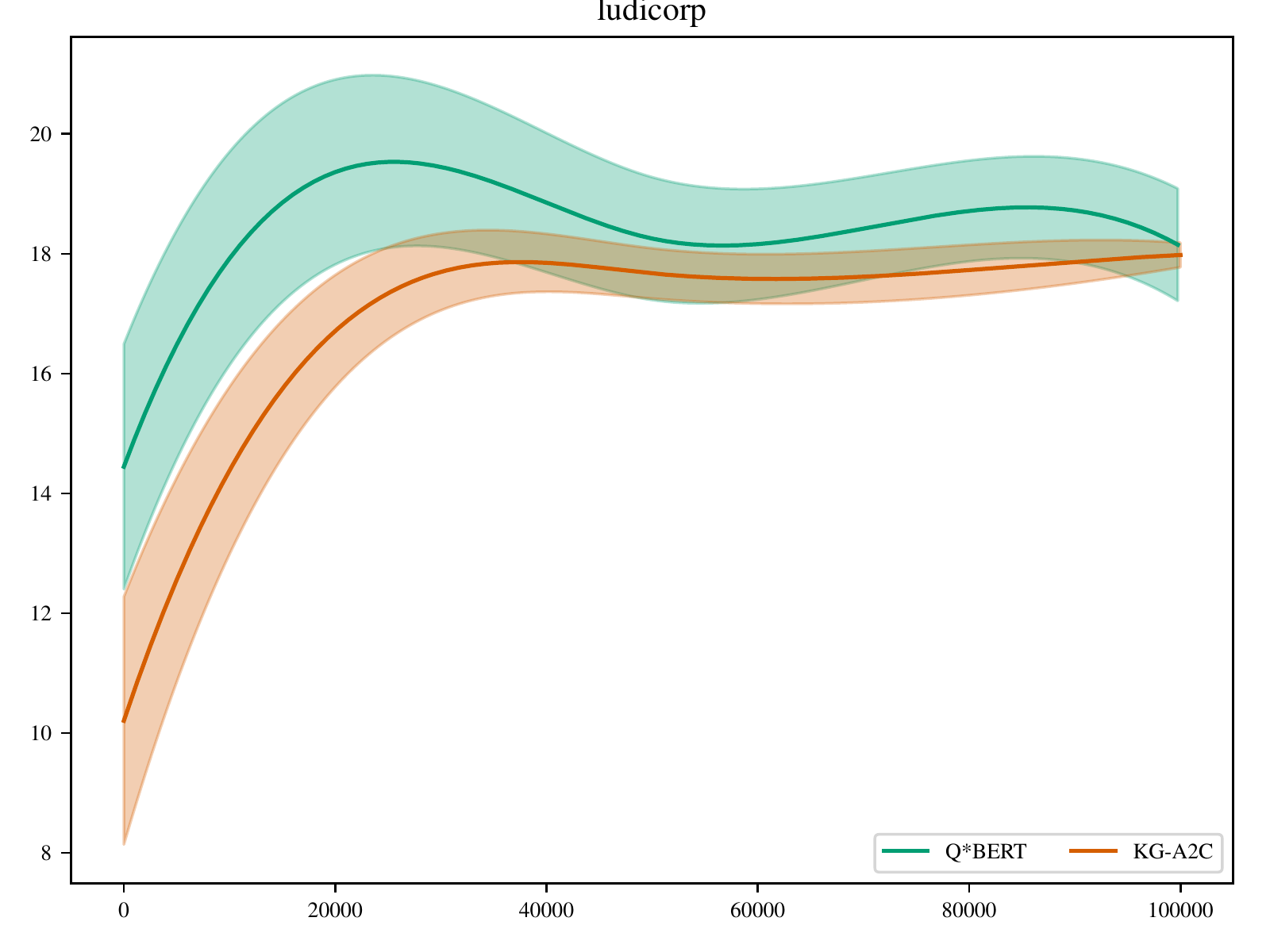}
\end{minipage}
\begin{minipage}{.33\textwidth}
  \centering
  \includegraphics[width=\linewidth]{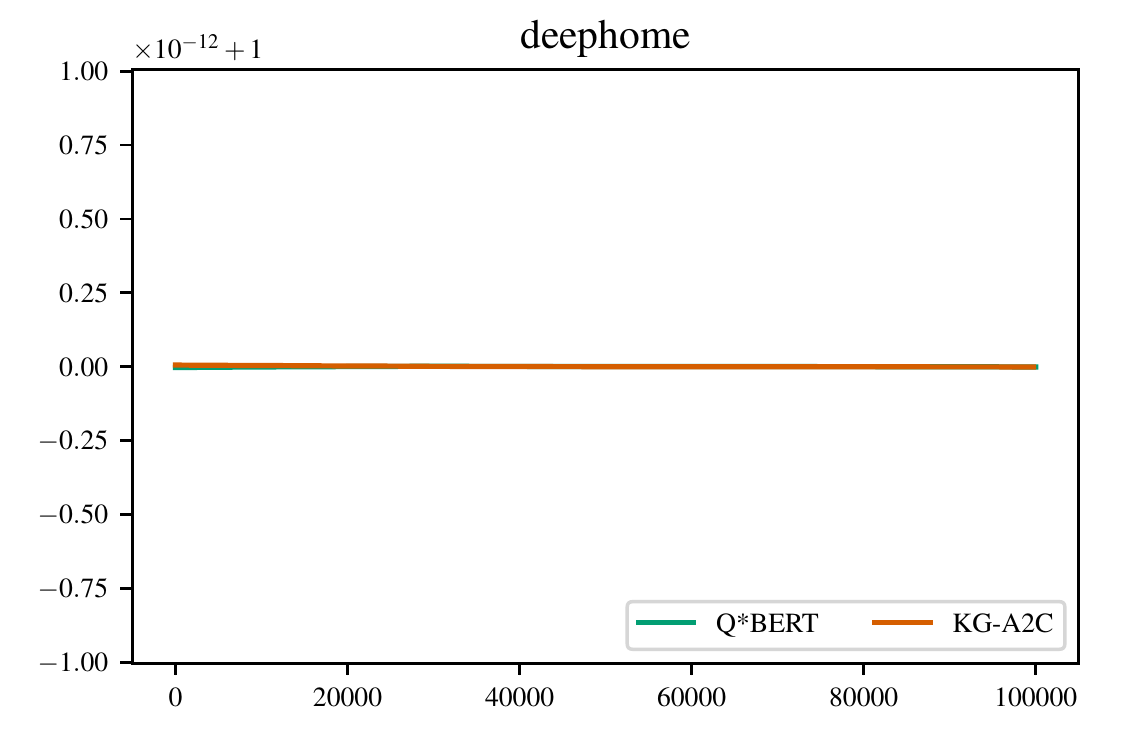}
\end{minipage}
\begin{minipage}{.33\textwidth}
  \centering
  \includegraphics[width=\linewidth]{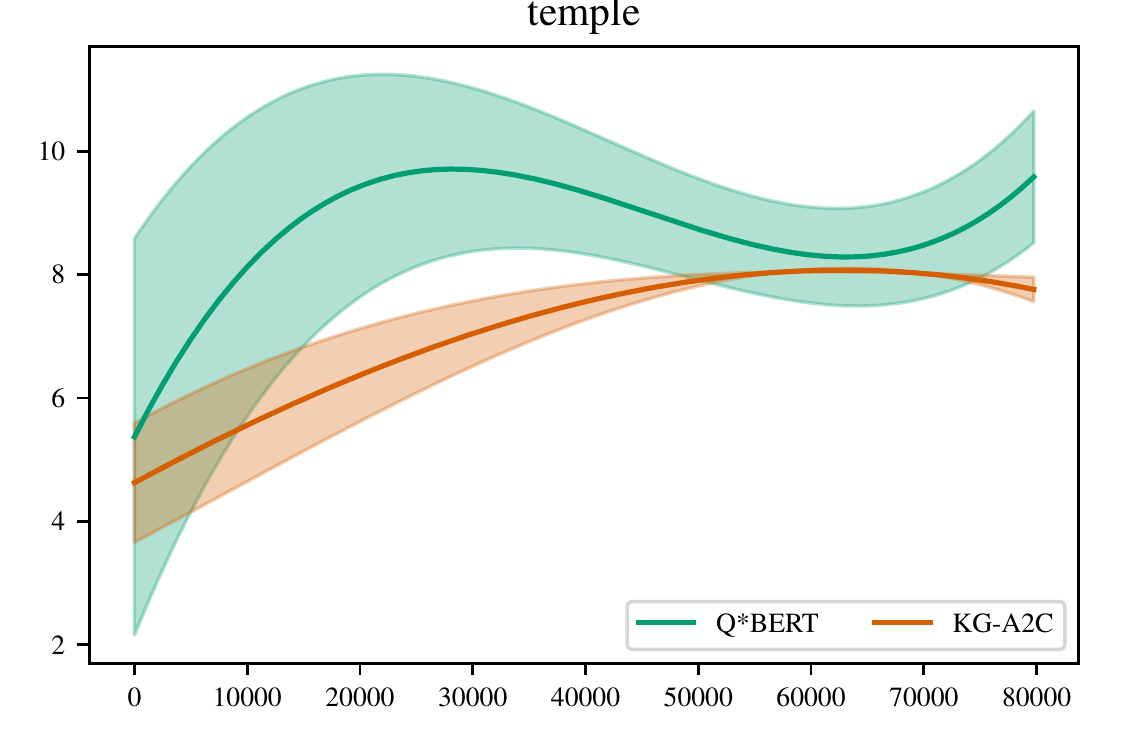}
\end{minipage}
\caption{Episode initial reward curves for KG-A2C and \qbert{}.}
\end{figure}

\newpage
\subsection{Intrinsic Motivation and Structured Exploration Results}
\begin{figure}[h]
\begin{minipage}{.33\textwidth}
  \centering
\includegraphics[width=\linewidth]{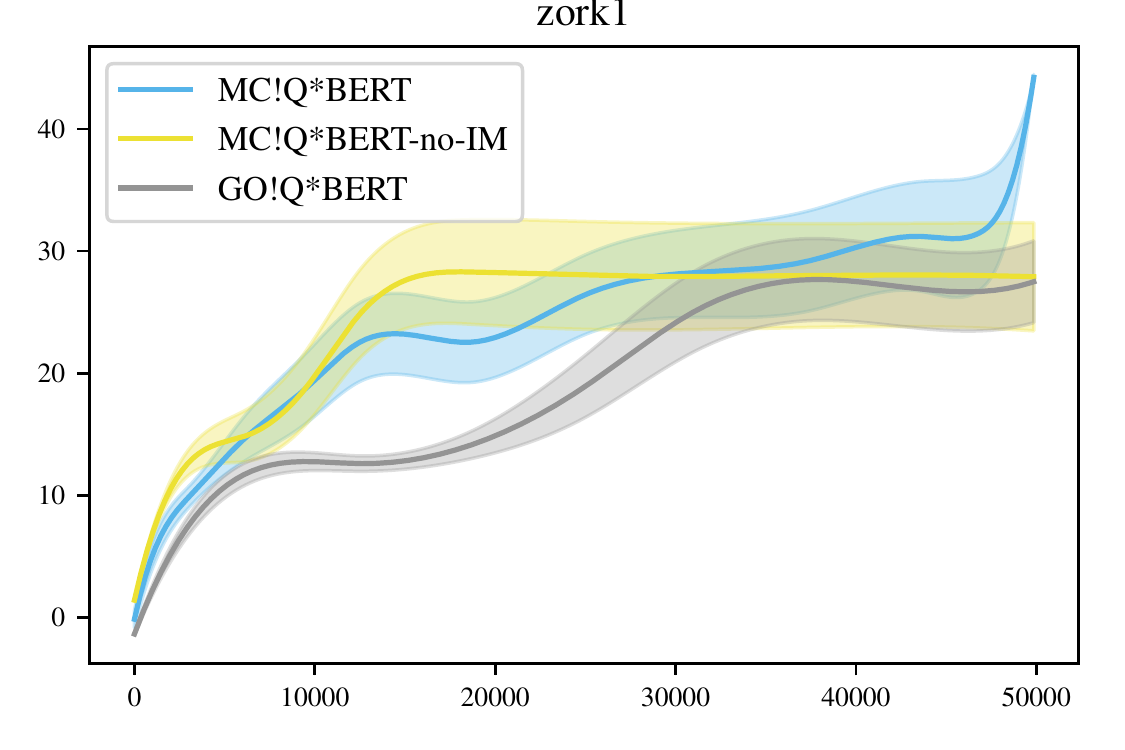}\end{minipage}
\begin{minipage}{.33\textwidth}
  \centering
  \includegraphics[width=\linewidth]{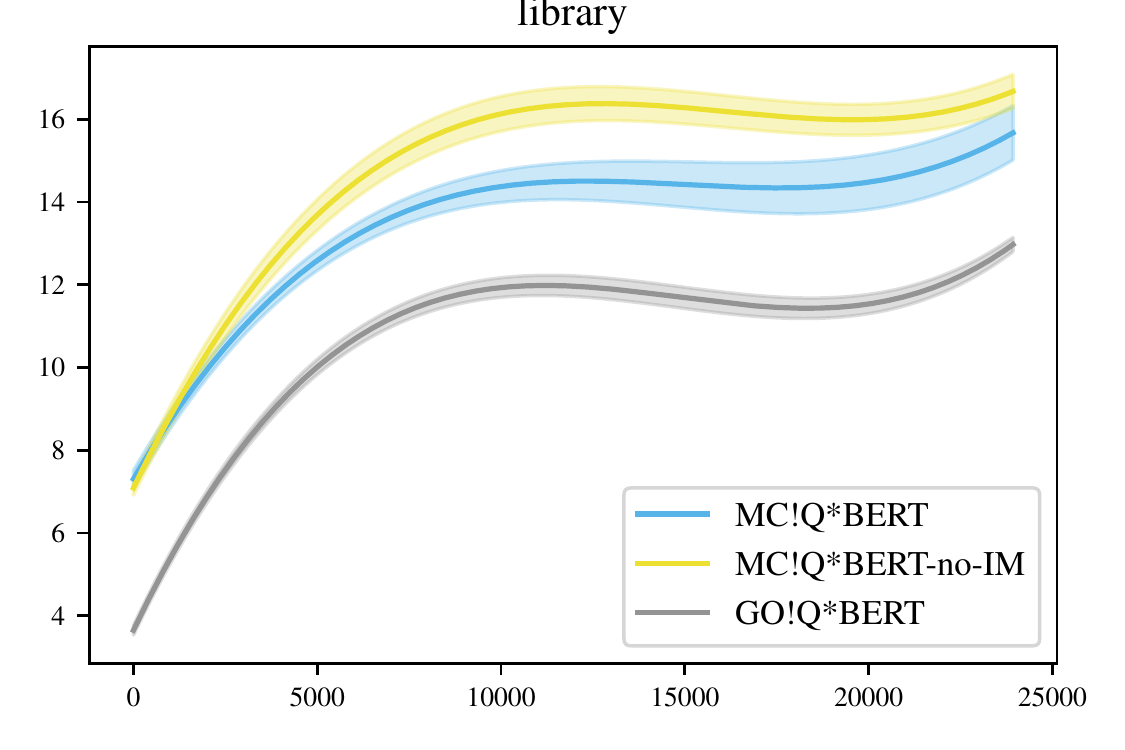}
\end{minipage}
\begin{minipage}{.33\textwidth}
  \centering
  \includegraphics[width=\linewidth]{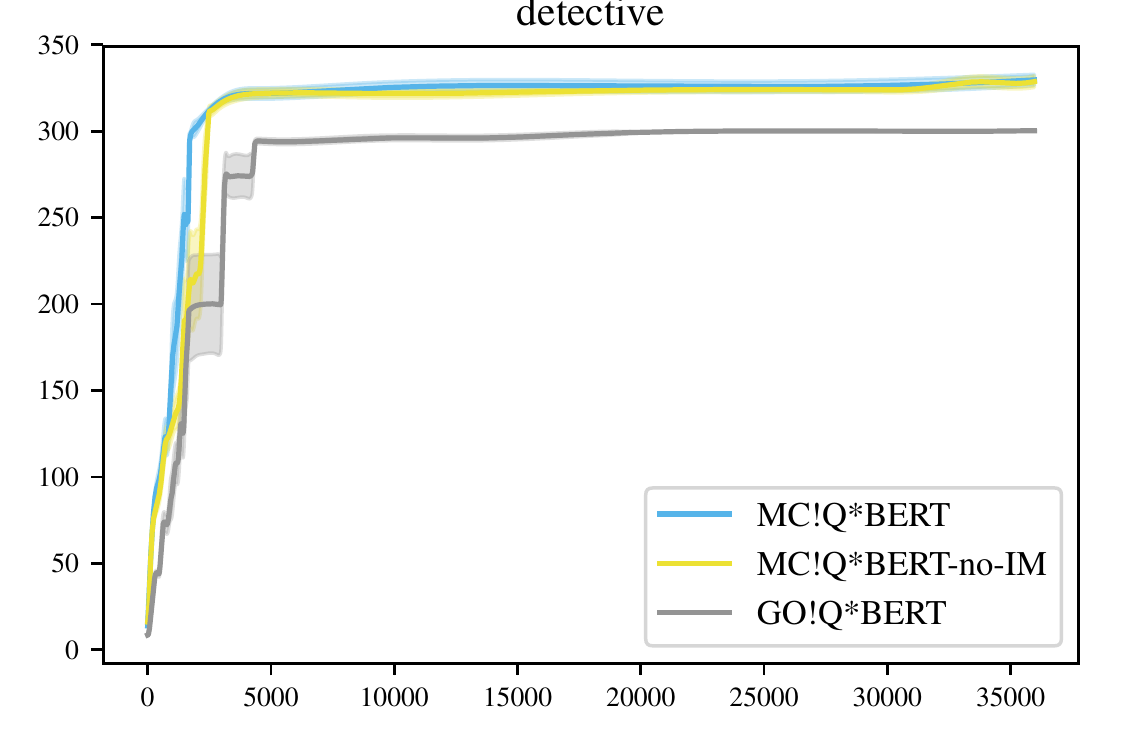}
\end{minipage}
\begin{minipage}{.33\textwidth}
  \centering
  \includegraphics[width=\linewidth]{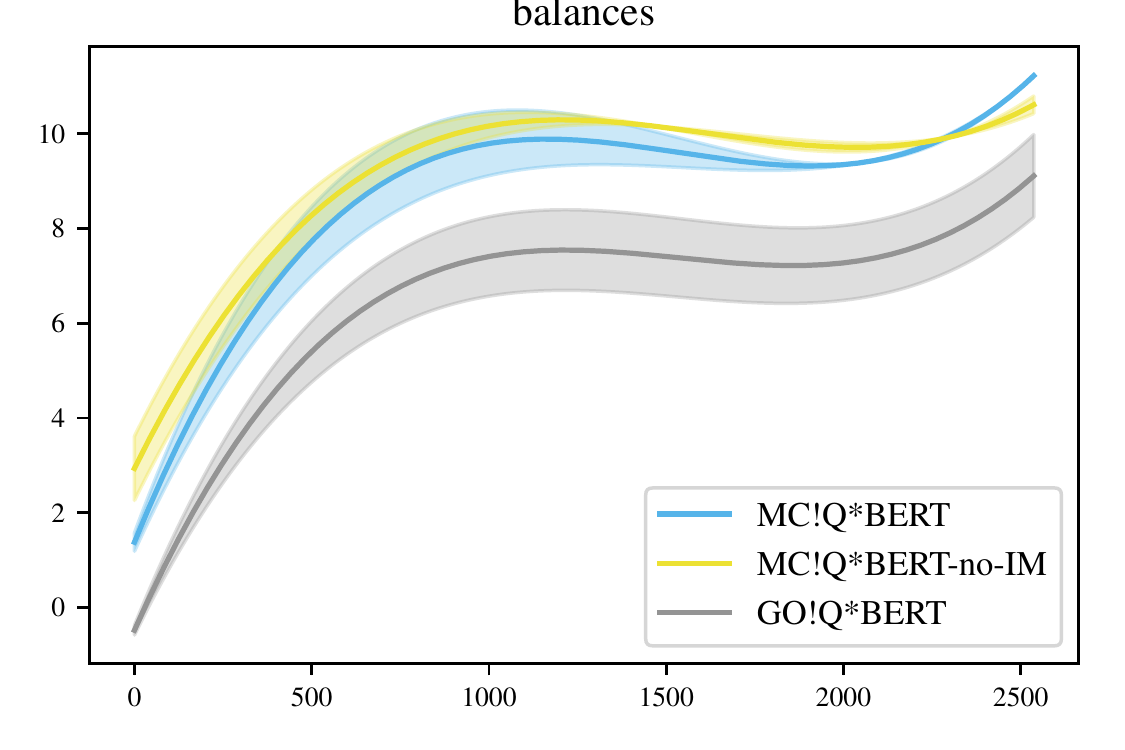}
\end{minipage}
\begin{minipage}{.33\textwidth}
  \centering
  \includegraphics[width=\linewidth]{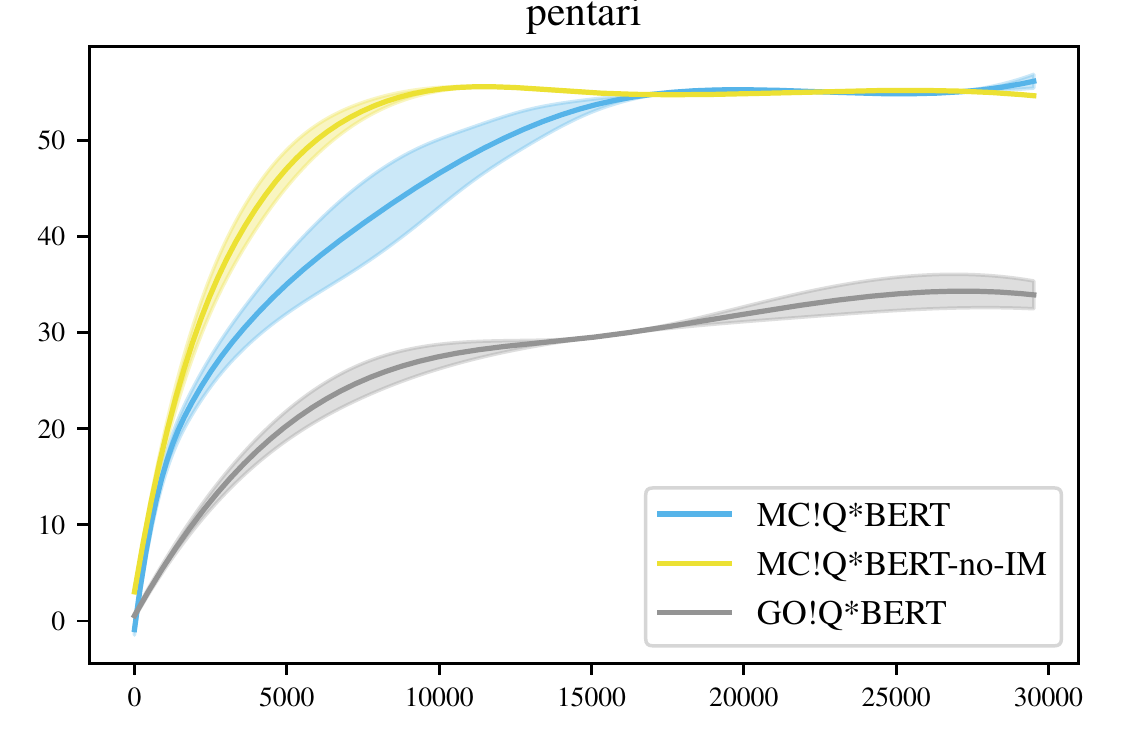}
\end{minipage}
\begin{minipage}{.33\textwidth}
  \centering
  \includegraphics[width=\linewidth]{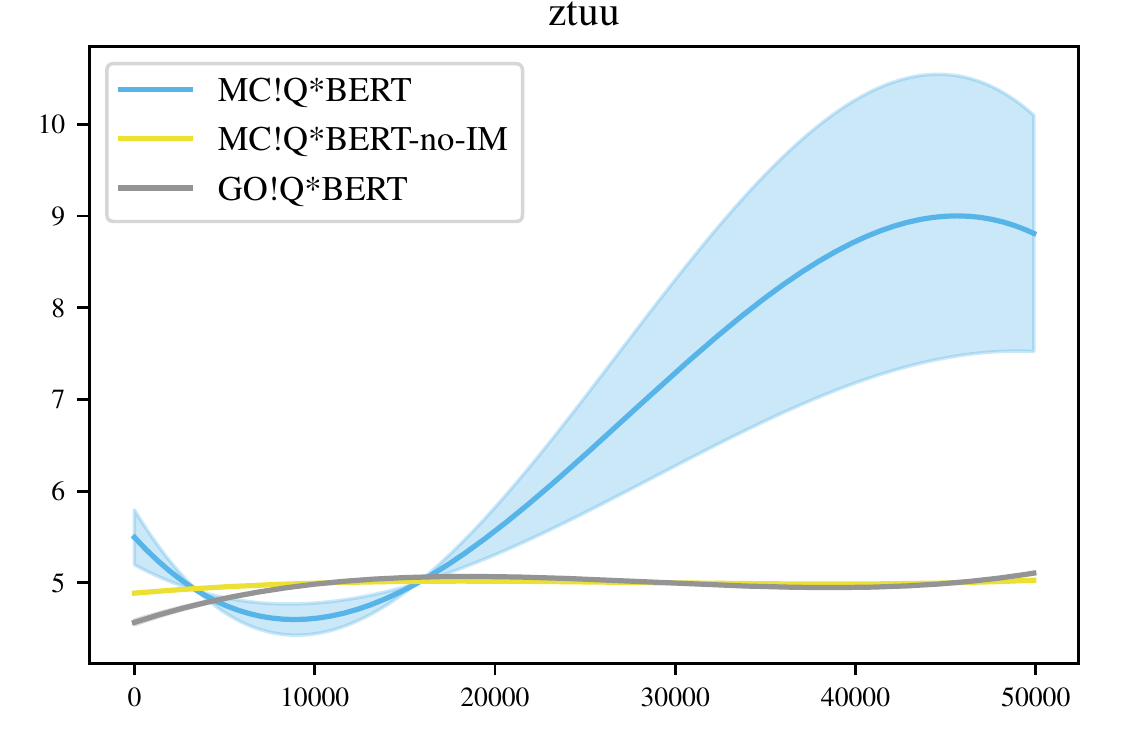}
\end{minipage}
\begin{minipage}{.33\textwidth}
  \centering
  \includegraphics[width=\linewidth]{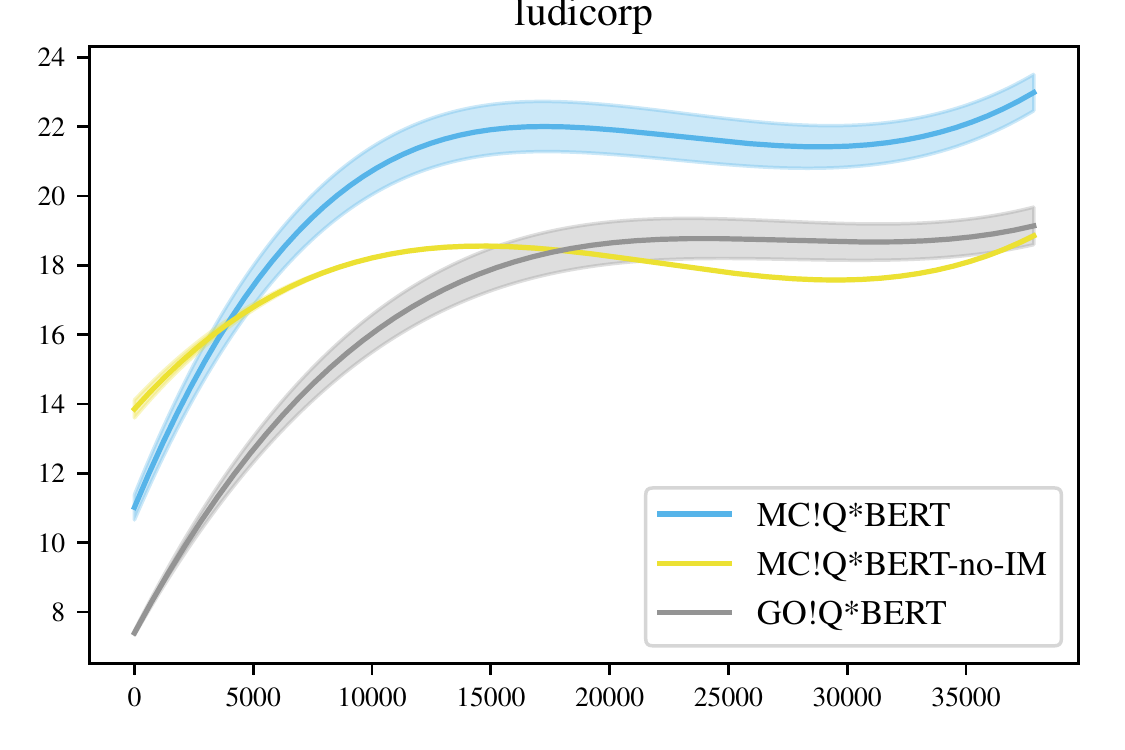}
\end{minipage}
\begin{minipage}{.33\textwidth}
  \centering
  \includegraphics[width=\linewidth]{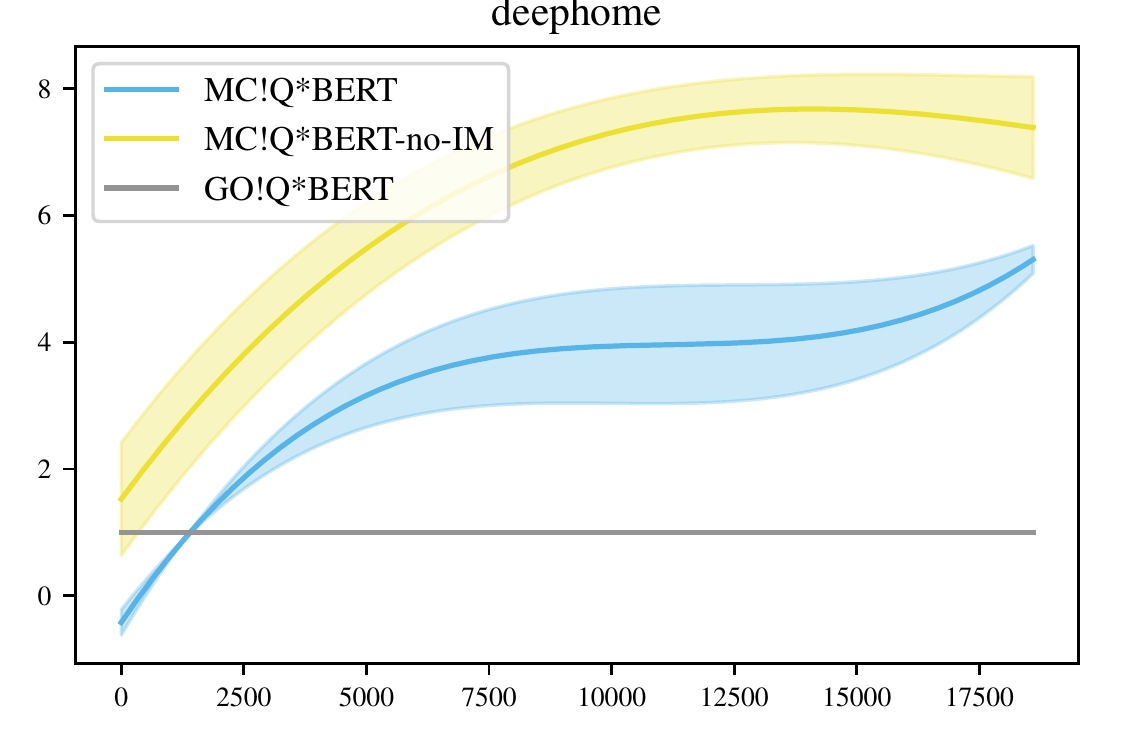}
\end{minipage}
\begin{minipage}{.33\textwidth}
  \centering
  \includegraphics[width=\linewidth]{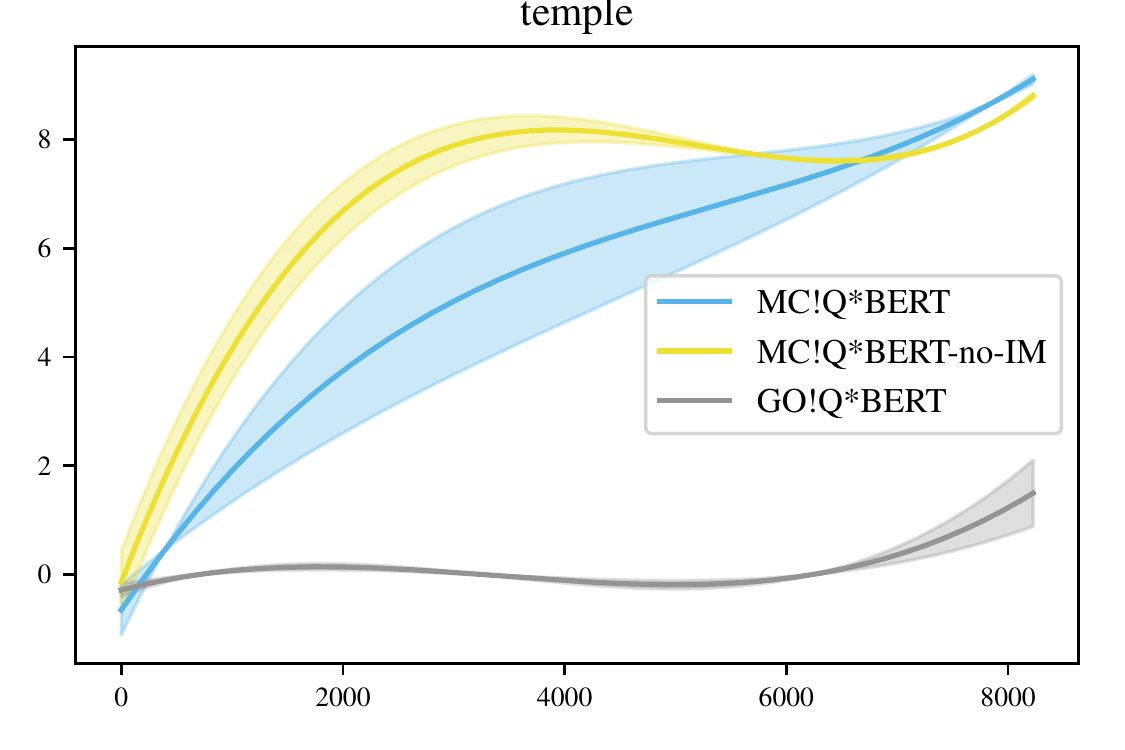}
\end{minipage}
\caption{Best initial reward curves for the exploration strategies.}
\end{figure}

\section{Zork1}

\end{document}